\documentclass{article} 
\usepackage{iclr2027_conference}

\usepackage{amsmath,amsfonts,bm}

\def\eqref#1{equation~\ref{#1}}

\def\1{\bm{1}}

\DeclareMathAlphabet{\mathsfit}{\encodingdefault}{\sfdefault}{m}{sl}
\SetMathAlphabet{\mathsfit}{bold}{\encodingdefault}{\sfdefault}{bx}{n}

\usepackage{bbm}
\usepackage{graphicx}
\usepackage{wrapfig}
\usepackage{capt-of}
\usepackage{hyperref}
\usepackage{url}
\usepackage{tabularx}
\usepackage{booktabs}
\usepackage{multirow}
\usepackage{graphicx}   
\usepackage{booktabs}   
\usepackage{multirow}   
\usepackage{array}      
\usepackage{makecell}   
\usepackage{amssymb}
\usepackage{pifont}
\usepackage[table]{xcolor}
\definecolor{best}{RGB}{254,236,221}
\definecolor{second}{RGB}{225,241,236}

\newcommand{\cmark}{\textcolor{green!60!black}{\ding{51}}}
\newcommand{\xmark}{\textcolor{red!70!black}{\ding{55}}}
\title{TRACE: Trajectory Representation and Consistency Estimation for AI-Generated Video Detection}

\author{
\textbf{Huangsen Cao}$^{1}$, \textbf{Hongkang Chu}$^{2}$, \textbf{Siyao Yu}$^{1}$, \textbf{Xin Ding}$^{3}$\\
\textbf{ Jianfeng Dong}$^{4}$, \textbf{Yongwei Wang}$^{1}$ \\
$^{1}$Zhejiang University \\
$^{2}$University of the Chinese Academy of Sciences \\
$^{3}$Nanjing University of Information Science and Technology \\
$^{4}$AZhejiang Gongshang University \\
\texttt{huangsen\_cao@zju.edu.cn}, \texttt{yongwei.wang@zju.edu.cn}
}

\iclrfinalcopy 
\begin{document}

\maketitle

\begin{abstract}
Recent advances in generative video models have enabled the synthesis of visually realistic content, posing significant challenges to synthetic video detection. Existing detectors often rely on appearance artifacts, semantic inconsistencies, and temporal patterns that may be generator-specific, limitating generalization to unseen synthesis models. We investigate whether responses to a pretrained generative model provide more transferable forensic cues. Our key observation is that real and AI-generated videos exhibit distinct \emph{velocity responses} under a pretrained Flow Matching video model. This distinction persists when different pretrained video-generation backbones are used as probes, suggesting that velocity responses offer transferable forensic signals beyond visual artificts. Motivated by this observation, we propose \textbf{TRACE} (\emph{\underline{T}rajectory \underline{R}epresentation \underline{a}nd \underline{C}onsistency \underline{E}stimation}), a generation-process-aware framework for AI-generated video detection. TRACE leverages a pretrained video DiT as a velocity-field probe to extract representations at multiple flow time points, and models cross-frame consistency through velocity differences between adjacent frames. We further introduce a \emph{Real-Centered Trajectory Optimization} objective that encourages generator-invariant representation learning. Extensive experiments on AIGVDBench demonstrate that TRACE generalizes effectively across diverse generators, substantially outperforming prior state-of-the-art methods on unseen open- and closed-source video generation models.
\end{abstract}

\section{Introduction}

Recent advances in generative visual intelligence have enabled increasingly realistic and diverse visual synthesis. Diffusion models~\citep{esser2024scaling,gu2022vector} and Flow-Matching based models ~\citep{lipman2022flow} have demonstrated strong capabilities for generating videos from text, images, and existing videos~\citep{ho2022video,peebles2023dit,lipman2023flow,wan2025wan}. As synthesized quality improves, detecting AI-generated content becomes more challenging. Meanwhile, evolving of generative architectures continuously alters the artifacts present in synthesized videos, making it difficult to learn forensic cues that can generalize to unseen generators. These challenges motivate exploring forensic cues tied to the generative process than only to the rendered appearance.

Most existing detectors analyze directly on rendered video content. Image-level methods exploit local pixel dependencies, frequency patterns, or transferable semantic representations~\citep{wang2020cnn,ojha2023towards,tan2024rethinking,chen2025dual}, while video-specific approaches further model cross-frame consistency, fine-grained spatiotemporal artifacts, or feature trajectories~\citep{ma2025decof,chen2026demamba,corvi2025seeing,interno2025restra,cui2026pvp}. Despite their effectiveness, these methods primarily characterize the final outputs rather than the responses induced by an underlying generative model. As a result, supervised detectors may learn appearance- or motion-based shortcuts that are correlated with the generators observed during training, limiting their ability to generalize to unseen video sources. The substantial performance variation across generation sources reported by AIGVDBench further highlights this challenge~\citep{ma2026your}.

\begin{wrapfigure}{r}{0.5\textwidth}
\vspace{-\intextsep}
\centering
\setlength{\abovecaptionskip}{4pt}
\includegraphics[width=\linewidth]{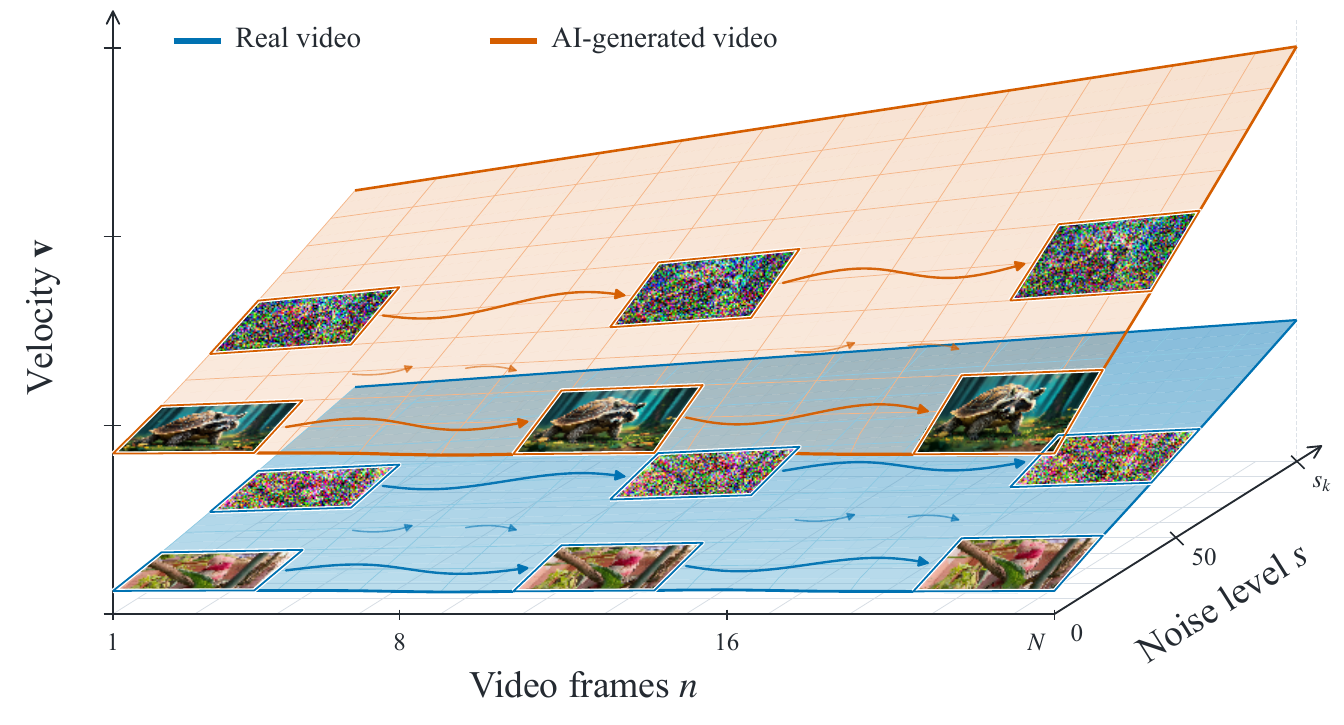}
\caption{Real videos exhibit smaller velocity responses than AI-generated videos, providing a discriminative cue for detection.}
\label{fig:velocity-response}
\vspace{-0.5\intextsep}
\end{wrapfigure}

A complementary direction is to exploit pretrained generative models as forensic probes. Existing image-based approaches investigate reconstruction errors or predicted-noise responses~\citep{wang2023dire,luo2024lare,zhang2023dnf}, while recent methods further explore denoising trajectories, probability-flow velocity, and curvature~\citep{vasilcoiu2025latte,liang2025denoising,jin2026dysydet,li2025curvature}. For video, related methods analyze diffusion reconstruction errors or score-derived spatiotemporal gradients~\citep{liu2024divid,chen2026reconfuse,zhang2025nsgvd}. However, it remains underexplored how to effectively represent videos through their responses to a pretrained Flow-Matching video DiT. In particular, the velocity field defines a time-dependent vector field that describes the transport between data and noise distributions~\citep{lipman2023flow}. Its responses to the same video at different flow times may therefore reveal generation-aware forensic cues that are complementary to those contained in the rendered content.

Motivated by this perspective, we investigate whether the generative dynamics encoded by a pretrained video generator can provide more transferable forensic evidence than rendered artifacts. Specifically, we probe the same video at multiple points along the generator's noising trajectory and examine its responses under the pretrained velocity field. As illustrated in Figure~\ref{fig:velocity-response}, real and AI-generated videos exhibit distinct velocity responses not only at individual flow times, but also in the variation of these responses along the generative trajectory. This motivates two complementary sources of evidence: flow-time-conditioned velocity responses and their local changes across neighboring flow times. Importantly, these changes occur along the generator's \emph{flow-time} dimension rather than the physical temporal dimension of the video, providing a distinct perspective from conventional inter-frame temporal-consistency analysis.

Based on this insight, we propose \textbf{TRACE}, which repurposes a pretrained video DiT as a generative probe to extract trajectory-aware representations from multiple flow times. TRACE further estimates the consistency of these representations across neighboring video frames by modeling their velocity differences. In addition, we observe that real videos generally exhibit smaller velocity responses than AI-generated videos, motivating a \emph{Real-Centered Trajectory Optimization} objective. This objective organizes the learned feature space around a real-video reference center by compacting real representations while pushing generated representations away from it. Such an asymmetric structure encourages the detector to characterize deviations from the real-video distribution rather than relying on generator-specific artifacts. By shifting forensic analysis from \emph{what is rendered} to \emph{how a pretrained generative prior responds}, TRACE aims to learn generation-aware representations with improved transferability across unseen synthesis sources.

Using Open-Sora as the sole synthetic training source, TRACE achieves macro-average AUCs of \textbf{97.95\%} and \textbf{95.11\%} on the open- and closed-source subsets of AIGVDBench, respectively. These results outperform the strongest baselines respectively by \textbf{12.87\%} and \textbf{17.01\%}, demonstrating the remarkable effectiveness of generative trajectory responses as transferable forensic signals for AI-generated video detection. TRACE further achieves the best average performance across six additional video-generation datasets, demonstrating strong cross-dataset generalization (Section~\ref{sec:additional-datasets}).

Our main contributions are threefold:
\begin{itemize}
\item We introduce a generation-aware forensic perspective for AI-generated video detection by probing a pretrained Flow-Matching video DiT at multiple flow times, revealing novel forensic cues beyond conventional rendered-content features.

\item We propose \textbf{TRACE}, which encodes multi-flow-time velocity responses, estimates cross-frame trajectory consistency through velocity differences, and proposes a real-centered optimization objective to facilitate generator-invariant representation learning.

\item We conduct extensive cross-generator evaluation on AIGVDBench, covering 31 generation settings spanning open- and closed-source models. TRACE achieves macro-average AUCs of \textbf{97.95\%} and \textbf{95.11\%}, outperforming the prior strongest baselines by \textbf{12.87\%} and \textbf{17.01\%}, respectively.
\end{itemize}

\section{Related Work}

\subsection{Video Generation Models}

Within diffusion-based \citep{esser2024scaling,gu2022vector,ho2020denoising} video synthesis, early models used spatiotemporal U-Nets \citep{ho2022video}, while recent systems scale latent video diffusion transformers \citep{peebles2023dit} for spatiotemporal modeling and text conditioning. Representative open models include Open-Sora \citep{zheng2024opensora} and its spatial--temporal DiT, e.g. CogVideoX \citep{yang2024cogvideox} with a 3D autoencoder and expert transformer, HunyuanVideo \citep{kong2024hunyuanvideo}, and the Wan family \citep{wan2025wan}. In parallel, Flow Matching \citep{lipman2023flow} learns continuous normalizing flows by regressing a time-dependent vector field along a prescribed probability path. The resulting velocity field describes how samples evolve between the data and noise distributions, rather than only the endpoint produced by a generator. Prior analysis has also shown that internal representations of video diffusion models contain motion-aware information \citep{xiao2024motion}. These findings suggest that a pretrained video generator can provide process-level representations for forensics, complementing cues measured only in rendered pixels.

\subsection{AI-Generated Visual Content Detection}

\textbf{Image-level detection.}
Early detectors learned transferable artifacts from CNN-generated images, with augmentation improving robustness to post-processing and unseen architectures~\citep{wang2020cnn}. UnivFD~\citep{ojha2023towards} leverages frozen CLIP~\citep{radford2021learning} features for cross-generator transfer, while NPR~\citep{tan2024rethinking}, D$^3$~\citep{yang2025d3}, and DDA~\citep{chen2025dual} target local dependencies and generator-invariant pixel-frequency cues. Recent image-forensics studies further explore adaptive feature modeling and explainable or interactive evidence analysis, including HyperDet~\citep{cao2027towards}, REVEAL~\citep{cao2025reveal}, Veritas~\citep{tan2026veritas}, and ClueAegis~\citep{cao2026clueaegis}. Beyond rendered-content analysis, another line of work exploits pretrained generative models as forensic probes through reconstruction errors, noise responses, and denoising trajectories, including DIRE~\citep{wang2023dire}, LaRE$^2$~\citep{luo2024lare}, DNF~\citep{zhang2023dnf}, TSG~\citep{zeng2024tsg}, timestep ensembling~\citep{wu2025timestep}, LATTE~\citep{vasilcoiu2025latte}, Denoising Trajectory Biases~\citep{liang2025denoising}, and DySy-Det~\citep{jin2026dysydet}. Recent studies further show that velocity and curvature can provide informative forensic cues~\citep{li2025curvature}. However, frame-wise image detection cannot fully exploit the temporal structure of videos.

\textbf{Video-level detection.}
Recent video detectors extend content-based analysis by modeling cross-frame consistency, spatiotemporal inconsistencies, low-level cues, or feature dynamics, including DeCoF \citep{ma2025decof}, DeMamba \citep{chen2026demamba}, WaveRep \citep{corvi2025seeing}, D3 \citep{zheng2025d3video}, ReStraV \citep{interno2025restra}, and V-PVP \citep{cui2026pvp}. STALL \citep{benhayun2026stall} and Skyra \citep{li2026skyra} further exploit real-video statistics and multimodal artifact reasoning, respectively. Meanwhile, process-based approaches extend generative-model probing to videos through diffusion reconstruction errors and score-derived spatiotemporal cues, as explored by DIVID \citep{liu2024divid}, ReConFuse \citep{chen2026reconfuse}, and NSG-VD \citep{zhang2025nsgvd}. Real-centered methods, such as Beyond Generation \citep{zhong2025beyond}, RCDN \citep{mccurdy2026rcdn}, LRD-Net \citep{zhang2026lrdnet}, and SphereVideo \citep{li2026spherevideo}, instead characterize authentic video distributions to improve detection robustness. Despite these advances, large-scale evaluation reveals substantial performance variation across detectors and generation sources~\citep{ma2026your}, while existing process-based video methods focus primarily on diffusion models. The forensic potential of multi-flow-time velocity responses from Flow-Matching video generators therefore remains largely unexplored. TRACE addresses this gap by modeling multi-step flow velocities and their cross-frame differences from a pretrained Flow-Matching video DiT.

\section{Methodology}

\begin{figure}[t]
\centering
\includegraphics[width=\textwidth]{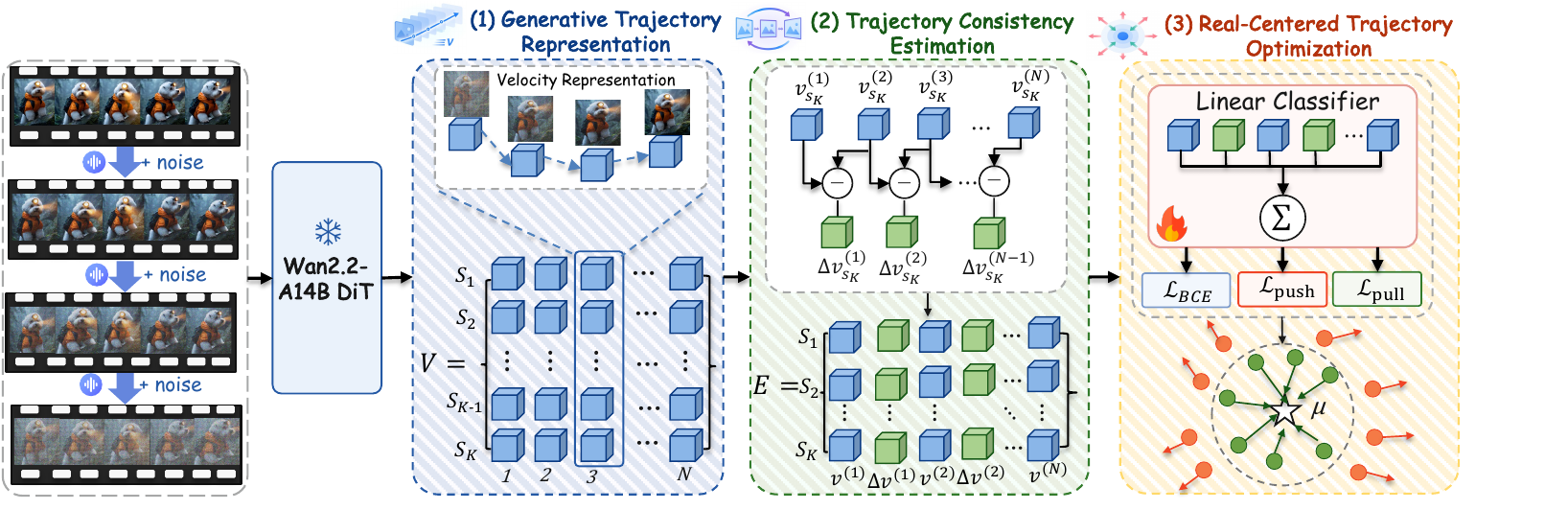}
\caption{\textbf{Overview of TRACE.} The proposed framework learns generative trajectory representations, estimates adjacent-frame consistency through velocity differences, and constrains the resulting feature space around a real-video center.}
\label{fig:trace-overview}
\vspace{-5mm}
\end{figure}

TRACE is motivated by a key observation that real and AI-generated videos exhibit distinct velocity responses in pretrained Flow Matching models. Based on this finding, TRACE models video generation dynamics beyond appearance-level artifacts. As illustrated in Figure~\ref{fig:velocity-response}, controlled noise perturbations reveal differences in these responses. The overall framework, shown in Figure~\ref{fig:trace-overview}, consists of three components: \emph{Generative Trajectory Representation Learning}, \emph{Trajectory Consistency Estimation}, and \emph{Real-Centered Trajectory Optimization}. Together, they learn generation-aware representations to improve generalization to unseen video generators.

\subsection{Generative Trajectory Representation Learning}

Recent Flow Matching-based video generation models formulate generation as a transport process between data and noise distributions, parameterized by a velocity field. TRACE exploits the responses of a pretrained velocity field to capture generation-aware cues beyond appearance-level artifacts.

Let $\mathbf{x}_0$ denote the clean latent of a video frame and $\boldsymbol{\epsilon}\sim p_0$ a noise sample. In the continuous Flow Matching formulation,
\begin{equation}
\mathbf{x}_t
=
(1-t)\mathbf{x}_0
+
t\boldsymbol{\epsilon},
\qquad
t\in[0,1].
\end{equation}

Rather than performing the complete generation or denoising process, TRACE treats the pretrained Wan2.2-A14B DiT as a fixed velocity field and probes it at a small set of scheduler timesteps. Let
\begin{equation}
\mathcal{S}
=
\{s_1,s_2,\ldots,s_K\}.
\end{equation}
Each $s_k$ specifies the noise level applied to the clean latent before the DiT forward pass, and thus represents a probing location rather than a denoising step. We use a 1000-level scheduler with $K=5$:
\begin{equation}
\mathcal{S}
=
\{0,25,50,75,100\}.
\end{equation}
These probe points are fixed throughout all experiments. As analyzed in Appendix~\ref{app:velocity_analysis}, $s=100$ already provides strong performance, while larger noise levels yield limited additional gains.


To capture temporal information, we uniformly sample $N$ frames from each video. Let $\mathbf{x}^{(n)}_0$ denote the clean latent of the $n$-th frame. A single noise realization $\boldsymbol{\epsilon}\sim p_0$ is shared across all sampled frames. For each $s_k\in\mathcal{S}$, the perturbed latent is obtained using the Wan2.2 Flow Matching scheduler:
\begin{equation}
\mathbf{x}^{(n)}_{s_k}
=
\mathcal{P}
\left(
\mathbf{x}^{(n)}_0,
\boldsymbol{\epsilon},
s_k
\right),
\qquad
n=1,\ldots,N,\quad
k=1,\ldots,K,
\end{equation}
where $\mathcal{P}(\cdot)$ denotes the scheduler-defined perturbation operation. At $s_k=0$, the clean latent is directly used.

The perturbed latent is then fed into the frozen DiT to obtain the velocity representation:
\begin{equation}
\mathbf{v}^{(n)}_{s_k}
=
f_{\theta}^{\mathrm{DiT}}
\left(
\mathbf{x}^{(n)}_{s_k},
s_k
\right).
\end{equation}
For each frame, the velocity responses across probe points form a generation-aware trajectory:
\begin{equation}
\mathcal{V}^{(n)}
=
\left[
\mathbf{v}^{(n)}_{s_1},
\mathbf{v}^{(n)}_{s_2},
\ldots,
\mathbf{v}^{(n)}_{s_K}
\right].
\end{equation}

Stacking the trajectories of all sampled frames yields the frame--noise representation:
\begin{equation}
\mathbf{V}
=
\begin{bmatrix}
\mathbf{v}^{(1)}_{s_1} &
\mathbf{v}^{(2)}_{s_1} &
\cdots &
\mathbf{v}^{(N)}_{s_1}
\\
\mathbf{v}^{(1)}_{s_2} &
\mathbf{v}^{(2)}_{s_2} &
\cdots &
\mathbf{v}^{(N)}_{s_2}
\\
\vdots &
\vdots &
\ddots &
\vdots
\\
\mathbf{v}^{(1)}_{s_K} &
\mathbf{v}^{(2)}_{s_K} &
\cdots &
\mathbf{v}^{(N)}_{s_K}
\end{bmatrix}
\in
\mathbb{R}^{K\times N\times D},
\end{equation}
where $D$ denotes the velocity feature dimension. The horizontal and vertical axes correspond to temporal progression and noise-level variation, respectively. Thus, $\mathbf{V}$ jointly preserves temporal and noise-conditioned responses of the pretrained generative model, providing generation-aware representations for subsequent trajectory consistency modeling and classification.

\subsection{Trajectory Consistency Estimation}

The velocity representation $\mathbf{V}$ characterizes the responses of individual video frames to the pretrained Flow Matching model at different scheduler timesteps. While these frame-level responses provide generation-aware cues, they do not explicitly capture temporal relationships between neighboring frames. We therefore introduce \emph{Trajectory Consistency Estimation} to model temporal consistency in the velocity space.

Neighboring frames in real videos typically describe the same underlying scene and evolve smoothly, leading to relatively consistent velocity responses. In contrast, temporal inconsistencies in AI-generated videos may induce irregular variations between adjacent responses. We thus measure frame-to-frame velocity changes while keeping the scheduler timestep and noise realization fixed. For the $k$-th probe timestep $s_k\in\mathcal{S}$, we compute
\begin{equation}
\Delta\mathbf{v}^{(n)}_{s_k}
=
\mathbf{v}^{(n+1)}_{s_k}
-
\mathbf{v}^{(n)}_{s_k},
\qquad
n=1,\ldots,N-1.
\end{equation}
Since the two frames share the same scheduler timestep and noise realization $\boldsymbol{\epsilon}$, the resulting difference primarily captures the change in model response induced by temporal variations in the video content. Repeating this operation across multiple probe timesteps yields
\begin{equation}
\Delta\mathbf{v}^{(1)}_{s_k},
\Delta\mathbf{v}^{(2)}_{s_k},
\ldots,
\Delta\mathbf{v}^{(N-1)}_{s_k},
\end{equation}
which characterizes temporal consistency under different noise conditions.

The original velocity representation $\mathbf{V}$ and temporal velocity differences provide complementary cues: $\mathbf{V}$ preserves the responses of individual frames, while $\Delta\mathbf{v}^{(n)}_{s_k}$ captures their temporal variations. We therefore fuse them into the final trajectory-aware representation:
\begin{equation}
\mathbf{E}
=
\operatorname{Fuse}
\left(
\mathbf{V},
\left\{
\Delta\mathbf{v}^{(n)}_{s_k}
\right\}
\right),
\end{equation}
where $\operatorname{Fuse}(\cdot)$ denotes the feature fusion operation and $\mathbf{E}$ represents the final trajectory-aware video feature. This representation jointly captures generation-aware velocity responses and temporal consistency cues, and is subsequently used for \emph{Real-Centered Trajectory Optimization}.

\subsection{Real-Centered Trajectory Optimization}

Although trajectory-aware representations capture generation-process and temporal cues, binary supervision may still encourage the detector to exploit generator-specific characteristics that are correlated with the training sources, limiting generalization to unseen generators. To address this issue, TRACE introduces a \emph{Real-Centered Trajectory Optimization} objective. Our observation that real videos generally exhibit smaller velocity responses than AI-generated videos motivates us to use real videos as the reference distribution. Specifically, TRACE encourages real-video representations to form a compact region around a real-feature center while pushing AI-generated representations away from this center. This asymmetric structure encourages the detector to learn deviations from the real-video distribution rather than relying on generator-specific artifacts.

Let $\mathbf{z}$ denote the fused trajectory representation and let $y\in\{0,1\}$ denote the ground-truth label, where $y=0$ represents a real video and $y=1$ represents an AI-generated video. The classification head produces a logit
\begin{equation}
\ell
=
h(\mathbf{z}),
\end{equation}
and the standard binary classification loss is defined as
\begin{equation}
\mathcal{L}_{\mathrm{BCE}}
=
\operatorname{BCEWithLogits}(\ell,y).
\end{equation}

To characterize the real-video distribution, we maintain a real-feature center $\boldsymbol{\mu}$ using an exponential moving average (EMA). Given the mean feature $\bar{\mathbf{z}}^{\,r}$ of a real batch, the center is updated by
\begin{equation}
\boldsymbol{\mu}
\leftarrow
\alpha\boldsymbol{\mu}
+
(1-\alpha)\bar{\mathbf{z}}^{\,r},
\end{equation}
where $\alpha=0.99$. We then measure the distance between an individual feature and the real center as
\begin{equation}
d
=
\left\|
\mathbf{z}
-
\boldsymbol{\mu}
\right\|_2.
\end{equation}

For real videos, we introduce a \emph{Pull} objective that encourages their trajectory-aware representations to concentrate around the real center:
\begin{equation}
\mathcal{L}_{\mathrm{pull}}
=
\lambda_{\mathrm{pull}}
\mathbbm{1}[y=0]
d^2,
\end{equation}
where $\lambda_{\mathrm{pull}}$ controls the strength of the real-centered constraint.

In contrast, AI-generated videos are encouraged to move away from the real center. We therefore introduce a \emph{Push} objective:
\begin{equation}
\mathcal{L}_{\mathrm{push}}
=
\lambda_{\mathrm{push}}
\mathbbm{1}[y=1]
\operatorname{softplus}
\left(
\frac{m-d}{m}
\right),
\end{equation}
where $m$ denotes the separation margin. This objective penalizes generated samples when their representations are close to the real center and gradually reduces the penalty as their distance exceeds the margin. Therefore, it does not impose an unnecessary constraint on already well-separated generated samples.

The overall training objective is
\begin{equation}
\mathcal{L}
=
w_{\mathrm{BCE}}\mathcal{L}_{\mathrm{BCE}}
+
\lambda_{\mathrm{pull}} \mathcal{L}_{\mathrm{pull}}
+
\lambda_{\mathrm{push}} \mathcal{L}_{\mathrm{push}},
\end{equation}
where $w_{\mathrm{BCE}}$, $\lambda_{\mathrm{pull}}$, and $\lambda_{\mathrm{push}}$ are set to $1$ by default, and the separation margin is set to $m=40$. When the real-feature center has not yet been initialized, only $\mathcal{L}_{\mathrm{BCE}}$ is applied.

The proposed objective explicitly imposes an asymmetric structure on the trajectory-aware feature space: real videos are pulled toward a compact real-centered region, whereas AI-generated videos are pushed away from it. This formulation encourages TRACE to focus on deviations from the real-video response pattern and reduces its reliance on generator-specific characteristics. As a result, the learned representation is better suited to generalize across unseen video generators. During inference, TRACE uses the classification logit $\ell$ as the detection score. The distance $d$ is used only during training to construct the real-centered optimization objective.

\section{Experiments}

\subsection{Experimental Setup}

To evaluate \textbf{TRACE}, we adopt \textit{Wan2.2-A14B}~\citep{wan2025wan} as the pretrained backbone and attach an MLP classifier to its final-layer DiT features. Experiments are conducted on \textit{AIGVDBench}~\citep{ma2026your}, the largest and most comprehensive benchmark for AI-generated video detection to date. Following a generator-disjoint evaluation protocol, TRACE is trained on a balanced subset of AIGVDBench comprising 14K real videos and 14K videos generated by Open-Sora, which serves as the sole synthetic source during training. For a fair comparison, all competing methods are trained on the same data. We evaluate the resulting detectors across 31 distinct generator--task configurations, covering 20 open-source and 11 closed-source settings, to assess their ability to generalize beyond the training generator. Detailed training and implementation configurations are provided in Appendix~\ref{app:training_detail}, detailed dataset information is provided in Appendix~\ref{app:dataset_detail}, and descriptions of the competing methods are given in Appendix~\ref{app:baseline_detail}.


\textbf{Evaluation Metrics.} We report \textbf{AUC} as the primary evaluation metric on AIGVDBench, measuring discrimination performance across decision thresholds. For completeness, AIGVDBench results for \textbf{ACC@5\% FPR} and \textbf{AP} are reported in Appendix~\ref{app:acc} and Appendix~\ref{app:ap}, respectively. We also use ACC@5\% FPR for the additional datasets in Section~\ref{sec:additional-datasets}. Visualizations of the experimental results are provided in Appendices~\ref{sec:Visualization} and~\ref{sec:t-sne}.


\subsection{Comparison to State-of-the-Art Detectors Evaluation on AIGVDbench}
We compare TRACE with state-of-the-art visual detectors, including both image- and video-based methods, on AIGVDBench. We report results separately on the open-source and closed-source generation settings to evaluate the cross-generator generalization of different detection paradigms.

\subsubsection{Results on Open-Source Data}

Table~\ref{tab:open-source-auc} reports the results on the open-source generator subset of AIGVDBench. Existing methods perform well primarily on in-domain Open-Sora videos but drop substantially on unseen generators. In contrast, TRACE achieves an AUC of \textbf{97.95\%}, outperforming the strongest baseline by \textbf{12.87\%}. This improvement indicates that modeling the \emph{generative trajectory} enables TRACE to capture generation-aware cues beyond generator-specific appearance artifacts, resulting in substantially stronger generalization to unseen generators.

\begin{table*}[htbp]
 \vspace{-5mm}
  \caption{\textbf{12.87\%} AUC improvement over existing detectors across open-source video generators, with the \colorbox{best}{\textbf{best}} and \colorbox{second}{\underline{second-best}} highlighted.}
  \resizebox{\textwidth}{!}{
    \begin{tabular}{>{\centering}m{2.5cm}|*{20}{c|}c}
    \toprule
    \multirow{3}{*}{\textbf{Method}} & \multicolumn{6}{c|}{\textbf{I2V}} & \multicolumn{12}{c|}{\textbf{T2V}} & \multicolumn{2}{c|}{\textbf{V2V}} & \multirow{3}{*}{\textbf{AVG}} \\
    \cmidrule(lr){2-7} \cmidrule(lr){8-19} \cmidrule(lr){20-21}
    & \multirow{2}{*}{\makecell[c]{\scriptsize Easy\\\scriptsize Animate}} & \multirow{2}{*}{\makecell[c]{\scriptsize LTX}} & \multirow{2}{*}{\makecell[c]{\scriptsize Pyramid\\\scriptsize Flow}} & \multirow{2}{*}{\makecell[c]{\scriptsize SEINE}} & \multirow{2}{*}{\makecell[c]{\scriptsize SVD}} & \multirow{2}{*}{\makecell[c]{\scriptsize Video\\\scriptsize Crafter}} & \multirow{2}{*}{\makecell[c]{\scriptsize Acc\\\scriptsize Video}} & \multirow{2}{*}{\makecell[c]{\scriptsize Animate\\\scriptsize Diff}} & \multirow{2}{*}{\makecell[c]{\scriptsize Cogvideo\\\scriptsize x1.5}} & \multirow{2}{*}{\makecell[c]{\scriptsize Easy\\\scriptsize Animate}} & \multirow{2}{*}{\makecell[c]{\scriptsize Hunyuan}} & \multirow{2}{*}{\makecell[c]{\scriptsize IPOC}} & \multirow{2}{*}{\makecell[c]{\scriptsize LTX}} & \multirow{2}{*}{\makecell[c]{\scriptsize Open\\\scriptsize Sora}} & \multirow{2}{*}{\makecell[c]{\scriptsize Pyramid\\\scriptsize Flow}} & \multirow{2}{*}{\makecell[c]{\scriptsize Rep\\\scriptsize Video}} & \multirow{2}{*}{\makecell[c]{\scriptsize Video\\\scriptsize Crafter}} & \multirow{2}{*}{\makecell[c]{\scriptsize Wan\\\scriptsize 2.1}} & \multirow{2}{*}{\makecell[c]{\scriptsize Cogvideo\\\scriptsize x1.5}} & \multirow{2}{*}{\makecell[c]{\scriptsize LTX}} & \\
    & & & & & & & & & & & & & & & & & & & & & \\

    \midrule
    
    \multicolumn{22}{c}{\textbf{AI-Generated Image Detection Models}} \\
\midrule

CNNSpot 
&76.71 &75.41 &89.59 &91.79 &97.72 &53.92 
&58.54 &39.90 &78.08 &64.15 &68.82 &85.43 
&95.72 &\cellcolor{best}\textbf{100.0} 
&96.37 &75.95 &72.78 &57.16 &71.42 &71.01 &76.02 \\

UnivFD
&76.39 &80.11 &87.65 &97.70 &95.79 &71.21
&61.32 &77.83 &87.53 &68.99 &60.68 &86.78
&89.47 &\cellcolor{second}\underline{99.99}
&94.24 &79.78 &87.59 &54.28 &81.94 &73.14 &80.62 \\

NPR
&77.27 &\cellcolor{second}\underline{88.64}
&\cellcolor{second}\underline{95.04} &93.58 &\cellcolor{second}\underline{98.07} &56.26
&42.96 &47.70 &78.34 &44.83 &57.74 &81.03
&\cellcolor{second}\underline{96.69} &\cellcolor{best}\textbf{100.0}
&93.63 &69.02 &72.69 &45.70 &\cellcolor{second}\underline{87.58} &84.71 &75.57 \\

DDA
&55.44 &52.00 &64.36 &62.55 &72.19 &40.60
&64.03 &67.84 &67.15 &73.67 &66.97 &58.09
&61.06 &\cellcolor{best}\textbf{100.0}
&83.45 &52.04 &62.72 &62.63 &52.92 &50.81 &63.53 \\

D3
&77.85 &82.16 &89.58 &\cellcolor{second}\underline{98.18}
&97.12 &82.20
&63.67 &72.67 &86.83 &73.80 &60.04 &90.09
&90.86 &\cellcolor{best}\textbf{100.0}
&96.25 &81.24 &90.51 &58.14 &81.09 &77.07 &82.47 \\

\midrule
\multicolumn{22}{c}{\textbf{AI-Generated Video Detection Models}} \\
\midrule

DeMamba
&70.08 &69.61 &80.82 &87.35 &94.50 &58.75
&62.80 &73.81 &79.05 &76.53 &71.46 &79.33
&83.77 &\cellcolor{best}\textbf{100.0}
&91.21 &66.99 &92.18 &62.01 &70.99 &66.06 &76.87 \\

DeCoF
&77.04 &81.92 &86.83 &93.66 &97.26 &60.24
&\cellcolor{second}\underline{78.46}
&73.99 &\cellcolor{second}\underline{91.23} &81.74 &\cellcolor{second}\underline{79.96}
&\cellcolor{second}\underline{95.29}
&93.14 &\cellcolor{best}\textbf{100.0}
&\cellcolor{second}\underline{98.51}
&90.38 &91.91 &68.01 &81.42 &\cellcolor{second}\underline{80.68}
&\cellcolor{second}\underline{85.08} \\

WaveRep
&49.74 &48.66 &49.40 &49.60 &49.03 &49.44
&48.89 &49.03 &50.04 &51.69 &46.37 &49.85
&50.66 &50.27
&51.28 &49.31 &52.63 &49.42 &46.86 &49.00 &49.56 \\

ReStraV
&\cellcolor{second}\underline{97.08}
&45.42 &49.48 &46.96 &60.89 &32.46
&42.23 &15.69 &73.32 &\cellcolor{second}\underline{91.25}
&46.01 &84.31
&47.22 &98.79
&44.32 &\cellcolor{second}\underline{91.36}
&15.28 &\cellcolor{second}\underline{76.89}
&54.08 &50.18 &58.16 \\

Qwen2.5-ViT
&51.14 &48.75 &53.85 &51.78 &50.69 &50.05
&56.67 &50.46 &60.53 &52.27 &56.84 &58.04
&52.30 &60.70
&55.85 &54.85 &55.25 &51.88 &57.07 &50.20 &53.96 \\

STALL & 50.29 & 46.47 & 47.41 & 45.12 & 50.05 & 54.02 & 56.89 & 67.25 & 55.52 & 62.23 & 59.58 & 62.68 & 46.17 & 60.21 & 56.15 & 54.88 & 64.06 & 57.93 & 49.95 & 48.52 & 54.77 \\

V-PVP
&72.16 &75.20 &84.17 &89.40 &93.16 &\cellcolor{second}\underline{83.54}
&75.23 &\cellcolor{second}\underline{85.64}
&76.56 &78.73 &76.30 &84.92
&81.75 &99.90
&93.07 &77.61 &\cellcolor{second}\underline{94.61}
&70.63 &72.47 &71.40 &81.82 \\

\textbf{\textit{TRACE}}
&
\cellcolor{best}\textbf{98.41}
&
\cellcolor{best}\textbf{99.06}
&
\cellcolor{best}\textbf{99.71}
&
\cellcolor{best}\textbf{100.0}
&
\cellcolor{best}\textbf{100.0}
&
\cellcolor{best}\textbf{100.0}
&
\cellcolor{best}\textbf{96.00}
&
\cellcolor{best}\textbf{99.95}
&
\cellcolor{best}\textbf{96.79}
&
\cellcolor{best}\textbf{99.23}
&
\cellcolor{best}\textbf{96.96}
&
\cellcolor{best}\textbf{98.77}
&
\cellcolor{best}\textbf{99.57}
&
\cellcolor{best}\textbf{100.0}
&
\cellcolor{best}\textbf{99.94}
&
\cellcolor{best}\textbf{95.91}
&
\cellcolor{best}\textbf{100.0}
&
\cellcolor{best}\textbf{84.11}
&
\cellcolor{best}\textbf{97.51}
&
\cellcolor{best}\textbf{97.12}
&
\cellcolor{best}\textbf{97.95}
\\
    
    \bottomrule
    \end{tabular}%
  }
  \label{tab:open-source-auc}%
  \vspace{-5mm}
\end{table*}%

\subsubsection{Results on Closed-Source Data}

\begin{table*}[htbp]
\centering

\caption{\textbf{17.01\%} AUC improvement over existing methods on closed-source video generators, with the \colorbox{best}{\textbf{best}} and \colorbox{second}{\underline{second-best}} highlighted.  }

\resizebox{\textwidth}{!}{
\begin{tabular}{
>{\centering\arraybackslash}m{3cm}|*{11}{c|}c
}

\toprule

\multirow{3}{*}{\textbf{Method}} 
& \multicolumn{11}{c|}{\textbf{Closed-Source Approaches}}
& \multirow{3}{*}{\textbf{AVG}} \\

\cmidrule(lr){2-12}
& \makecell[c]{\scriptsize Gen2}
& \makecell[c]{\scriptsize Gen3}
& \makecell[c]{\scriptsize Jimeng}
& \makecell[c]{\scriptsize Luma}
& \makecell[c]{\scriptsize Open\\\scriptsize Sora}
& \makecell[c]{\scriptsize Sora}
& \makecell[c]{\scriptsize Causvid\\\scriptsize 24fps}
& \makecell[c]{\scriptsize Kling}
& \makecell[c]{\scriptsize Pika}
& \makecell[c]{\scriptsize Vidu}
& \makecell[c]{\scriptsize Wan}

\\

\midrule

\multicolumn{13}{c}{\textbf{AI-Generated Image Detection Models}} \\

\midrule
CNNSpot &37.23 & 60.95 & 15.72 & 57.38 & 61.64 & 59.71 & 69.60 & 52.88 & 63.08 & 75.54 & 54.47 & 55.29 \\
UnivFD & 41.04 & 46.20 & 23.89 & 51.35 & 75.17 & 52.65 & 82.24 & 61.47 & 54.27 & 72.59 & 54.52 & 55.94 \\
NPR & 24.70 & 47.85 & 3.19 & 50.51 & 61.95 & 50.37 & 38.31 & 50.70 & 63.16 & 70.34 & 43.99 & 45.92 \\
DDA & 65.97 & 72.90 & 64.86 & 69.54 & 83.44 & 72.88 & 69.89 & 71.22 & 65.97 & 73.80 & 69.41 & 70.90 \\
D3 & 36.59 & 41.77 & 31.97 & 42.34 & 67.67 & 42.72 & 82.57 & 60.92 & 51.82 & 66.09 & 50.27 & 52.25 \\

\midrule
\multicolumn{13}{c}{\textbf{AI-Generated Video Detection Models}} \\

\midrule

DeMamba &68.57 & 64.46 & 54.10 & 61.67 & 78.56 & 66.44 & 68.63 & 54.98 & 76.26 & 72.54 & 57.22 & 65.77 \\
DeCoF& 54.31 & 69.45 & 29.02 & 73.18 & \cellcolor{second}\underline{90.69} & 74.85 & \cellcolor{second}\underline{95.48} & \cellcolor{second}\underline{74.54} & 83.10 & \cellcolor{second}\underline{84.41} & 70.82 & 72.71 \\
WaveRep & 56.27 & 53.20 & 52.56 & 51.44 & 55.20 & 57.26 & 54.78 & 48.16 & 53.04 & 55.04 & 52.12 & 53.55 \\
ReStraV  & 20.84 & 41.24 & 16.78 & 37.34 & 41.55 & 57.55 & 42.21 & 43.14 & 39.20 & 45.32 & 54.90 & 40.01 \\
Qwen2.5-ViT & 71.92 & 57.40 & 71.19 & 55.52 & 63.76 & 53.06 & 57.36 & 50.42 & 58.53 & 54.64 & 53.06 & 58.80 \\
STALL & \cellcolor{second}\underline{85.48} & \cellcolor{second}\underline{79.22} & \cellcolor{second}\underline{77.89} & \cellcolor{second}\underline{81.43} & 82.27 & \cellcolor{second}\underline{74.98} & 80.56 & 72.10 & 79.80 & 70.74 & \cellcolor{second}\underline{74.68} & \cellcolor{second}\underline{78.10} \\
V-PVP & 74.35 & 76.28 & 63.22 & 69.80 & 87.74 & 70.55 & 71.99 & 72.92 & \cellcolor{second}\underline{90.23} & 77.55 & 70.66 & 75.03 \\

\textbf{\textit{TRACE} } & \cellcolor{best}\textbf{99.48} & \cellcolor{best}\textbf{93.11} & \cellcolor{best}\textbf{99.50} & \cellcolor{best}\textbf{97.20} & \cellcolor{best}\textbf{99.38} & \cellcolor{best}\textbf{81.52} & \cellcolor{best}\textbf{97.94} & \cellcolor{best}\textbf{97.47} & \cellcolor{best}\textbf{99.86} & \cellcolor{best}\textbf{92.98} & \cellcolor{best}\textbf{87.80} & \cellcolor{best}\textbf{95.11} \\

\bottomrule

\end{tabular}
}

\label{tab:closed_source_auc}
\vspace{-5mm}
\end{table*}

Table~\ref{tab:closed_source_auc} reports the results on the closed-source generator subset of AIGVDBench. Compared with open-source generators, closed-source models present a more challenging setting, where existing methods suffer substantial performance degradation on unseen generators. In contrast, TRACE achieves an AUC of \textbf{95.11\%}, outperforming the strongest baseline by \textbf{17.01\%}. Notably, TRACE maintains strong performance across unseen closed-source generators, demonstrating its superior capability in cross-generator generalization.

\subsection{Evaluation on Additional Datasets}
\label{sec:additional-datasets}

\noindent
\begin{minipage}[t]{0.47\textwidth}
\vspace{0pt}
To assess generalization beyond AIGVDBench, we evaluate TRACE on six additional datasets: \textit{VideoFeedback}~\citep{he2024videoscore}, \textit{GenVideo}~\citep{chen2026demamba}, \textit{ComGenVid}~\citep{ben2026training}, \textit{Magic Videos}~\citep{li2026preserving}, \textit{GVD}~\citep{bai2024ai}, and \textit{GVF}~\citep{ma2025detecting}. Figure~\ref{fig:additional-datasets-radar} compares TRACE with DDA, DeMamba, DeCoF, STALL, and V-PVP using ACC@5\% FPR. TRACE ranks first on four of the six datasets, with an unweighted mean of \textbf{85.48\%}, compared with \textbf{76.48\%} for V-PVP, the strongest baseline on average. The largest gain over the best baseline is \textbf{15.60\%} on ComGenVid. STALL performs best on GenVideo and Magic Videos, indicating that the gains vary across datasets. While TRACE does 
\end{minipage}\hfill
\begin{minipage}[t]{0.49\textwidth}
\vspace{0pt}
\centering
\setlength{\parskip}{0pt}
\includegraphics[width=0.90\linewidth,trim=0 15 0 5,clip]{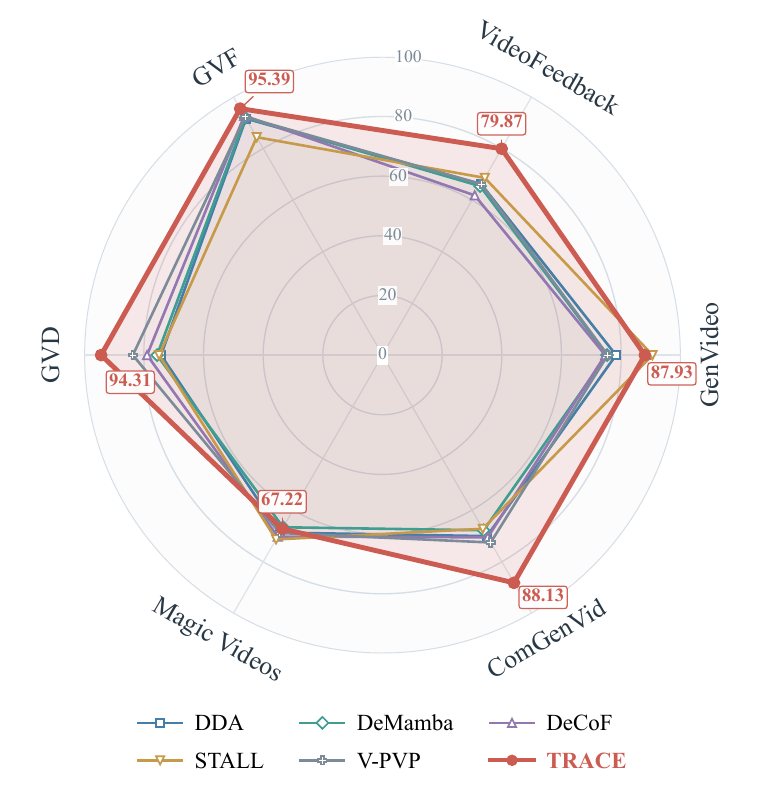}\par
\setlength{\abovecaptionskip}{0pt}
\setlength{\belowcaptionskip}{0pt}
\captionof{figure}{ACC@5\% FPR (\%) on six datasets.}
\label{fig:additional-datasets-radar}
\end{minipage}

\vspace{-2mm}
\par
not consistently outperform all baselines on every dataset, its performance remains comparable to that of existing state-of-the-art methods, demonstrating strong generalization across diverse video generation scenarios.
\subsection{Ablation Studies}

\subsubsection{Ablation of Individual Components}

We conduct ablation studies on the main components of TRACE to evaluate their individual contributions to the overall detection performance. As shown in Table~\ref{tab:component_ablation}, we investigate the effects of noise perturbation and Real-Centered Trajectory Optimization (RCTO), while keeping the remaining settings unchanged. The results demonstrate that each component contributes to the final performance, and their combination yields the most effective detection capability.

\subsubsection{Backbone Comparison}

We investigate the effectiveness of different pretrained video generation backbones for extracting generation-aware representations. Specifically, we replace the Wan2.2-A14B DiT backbone with \textit{HunyuanVideo-I2V}\citep{kong2024hunyuanvideo}, \textit{Wan2.1-1.3B}, \textit{Wan2.1-14B}, and \textit{Open-Sora}\citep{zheng2024open}, while keeping all other components of TRACE unchanged. As shown in Figure~\ref{fig:backbone-comparison}, the detection performance generally improves with the capability and scale of the pretrained video generation backbone, with the more advanced and larger models yielding stronger generation-aware representations. Meanwhile, all evaluated backbones provide effective velocity representations for distinguishing real and AI-generated videos, demonstrating that the velocity responses of pretrained generative models contain useful generation-related cues for video detection.

\begin{table}[t]
    \centering
    \begin{minipage}[t]{0.48\textwidth}
    \vspace{0pt}
    \centering
    \caption{Ablation study of TRACE components. ``Open'' and ``Closed'' denote open- and closed-source test sets, respectively.}
    \label{tab:component_ablation}
    \setlength{\tabcolsep}{3pt}
    \renewcommand{\arraystretch}{1.2}
    \resizebox{\linewidth}{!}{
    \begin{tabular}{cc|c|cc|cc}
        \toprule
        \multicolumn{2}{c|}{Components} &
        \multicolumn{1}{c|}{\makecell{Noise\\Level}} &
        \multicolumn{2}{c|}{AUC (\%)} &
        \multicolumn{2}{c}{ACC@5\%FPR (\%)} \\
        \cmidrule(lr){1-2}
        Noise & RCTO & $t$ &
        Open & Closed &
        Open & Closed \\
        \midrule
        \xmark & \xmark & -- &
        96.29& 91.70 &
        90.98 & 87.04 \\

        \cmark & \xmark & 100 &
        97.39 & 93.95 &
        93.09 & 90.69 \\

        \xmark & \cmark & -- &
        
        96.71 & 92.38 &
        91.54 & 87.78 \\

        \cmark & \cmark & 200 &
        97.21 & 93.86 &
        92.74 & 90.32 \\

        \cmark & \cmark & 100 &
        \textbf{97.95} & \textbf{95.11} &
        \textbf{94.27} & \textbf{92.03} \\
        \bottomrule
    \end{tabular}
    }
    \end{minipage}\hfill
    \begin{minipage}[t]{0.48\textwidth}
    \vspace{0pt}
    \centering
    \includegraphics[width=\linewidth]{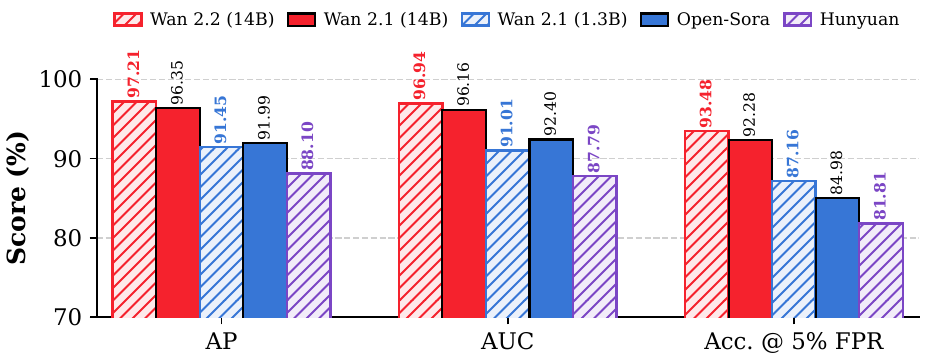}
    \captionof{figure}{Comparisons of different backbones on AIGVDBench. Reported results are averaged across all 31 settings.}
    \label{fig:backbone-comparison}
    \end{minipage}
    \vspace{-4mm}
\end{table}

\begin{figure}[htbp]
\centering
\includegraphics[width=\textwidth]{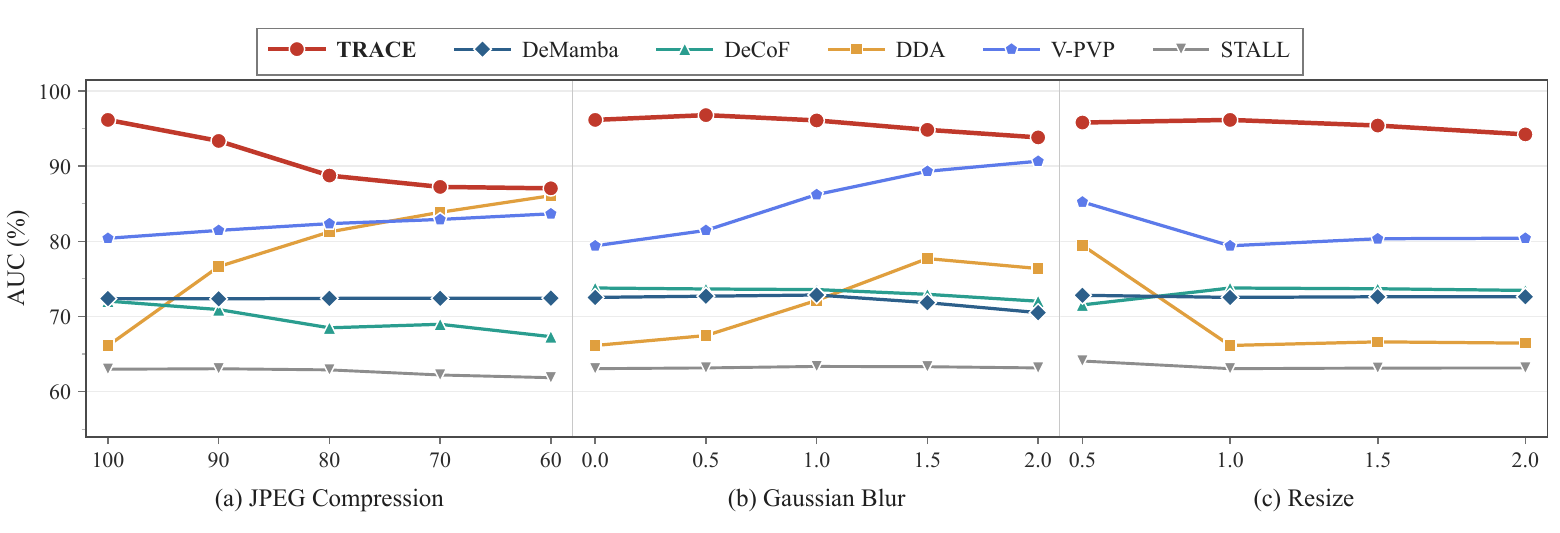}
\caption{Robustness evaluation of TRACE in terms of AUC under different video distortions, including JPEG compression, Gaussian blur, and resizing.}
\label{fig:robustness_AUC}
\end{figure}

\subsection{Robustness Evaluation}
We further evaluate the robustness of TRACE on AIGVDBench under JPEG compression ($q\in\{100,90,80,70,60\}$), Gaussian blur ($\sigma\in\{0.0,0.5,1.0,1.5,2.0\}$), and resolution scaling ($s\in\{0.5,1.0,1.5,2.0\}$, with restoration to the original size). Perturbations are applied to raw frames prior to model-specific preprocessing. We report AUC across all 14 settings, as shown in Figure~\ref{fig:robustness_AUC}. TRACE consistently achieves the best performance across the evaluated robustness conditions, demonstrating its strong robustness to common video distortions. This robustness can be attributed to its reliance on generation-aware velocity representations and trajectory-level consistency, which provide complementary forensic cues beyond appearance-based artifacts and help preserve the discriminative information under various input distortions.

\section{Conclusion}

This work presented \textbf{TRACE}, a trajectory-based framework for AI-generated video detection that extracts generation-aware cues from pretrained Flow Matching models. By modeling velocity responses across multiple flow time points and their cross-frame consistency, TRACE captures transferable forensic signals beyond conventional appearance-based artifacts. Extensive experiments demonstrate that TRACE significantly outperforms existing detectors in generalizing to unseen open- and closed-source video generators.



\bibliography{iclr2027_conference}
\bibliographystyle{iclr2027_conference}
\appendix
\clearpage


\section{Implementation and Training Details}
\label{app:training_detail}

\paragraph{Training Configuration.}
Unless otherwise specified, we freeze the VAE, T5 text encoder, and DiT backbone of Wan2.2-A14B, and optimize only the trajectory classification head. We use the low-noise DiT expert to probe each video at five flow-time points, $\mathcal{T}=\{0,25,50,75,100\}$. The resulting features are fused by concatenating the absolute probe features, adjacent flow-time differences along the noise path, and two temporal differences between clean-frame representations, corresponding to the middle--first and last--middle frame pairs. The resulting representation is fed into a lightweight classification head consisting of LayerNorm, Dropout with a rate of $0.1$, and a linear layer with one output unit. Following the default text-conditioning interface of Wan2.2, all video clips are associated with the fixed prompt ``this is a video.''

\paragraph{Preprocessing and Augmentation.}
For each video, we uniformly sample $9$ frames and process them at a spatial resolution of $480{\times}832$. We employ an adaptive crop-resize strategy: videos smaller than the target resolution are resized using bilinear interpolation, whereas larger videos are center-cropped without downscaling. During training, spatial appearance augmentation is applied with probability $0.5$, where a random subset of Gaussian blur, additive noise, and color jitter is sampled. Otherwise, the original frames are retained. In addition, a random temporal crop is independently applied with probability $0.5$ when the source video contains sufficient frames. All augmentations are disabled during evaluation, while the same spatial preprocessing is retained.

\paragraph{Objective and Optimization.}
The trajectory classifier is optimized with the following objective:
\begin{equation}
\mathcal{L}
=
\mathcal{L}_{\mathrm{BCE}}
+
\mathcal{L}_{\mathrm{pull}}
+
\mathcal{L}_{\mathrm{push}},
\end{equation}
where all three terms are assigned unit weights. $\mathcal{L}_{\mathrm{BCE}}$ denotes the binary cross-entropy loss. $\mathcal{L}_{\mathrm{pull}}$ encourages real-video features to remain close to an EMA-updated real-feature center with a decay rate of $0.99$, while $\mathcal{L}_{\mathrm{push}}$ encourages fake-video features to remain beyond a margin of $m=40$ from the real-feature center. We optimize the classifier using AdamW with a learning rate of $1{\times}10^{-4}$ and a weight decay of $0.01$. Training is performed with a per-GPU batch size of $4$ on $4$ GPUs, resulting in an effective batch size of $16$, for a total of $5$ epochs ($8{,}750$ optimization steps). Unless otherwise stated, all reported results are obtained from the checkpoint at step $8{,}750$. The frozen Wan2.2-A14B backbone is executed in bfloat16 precision.

\paragraph{Metric Computation.}
We report AUC as the primary threshold-independent metric. To avoid relying on an arbitrarily selected classification threshold when computing accuracy, we additionally report ACC@5\% FPR. Specifically, for each evaluation split, the decision threshold is determined as the $95$th percentile of the real-video scores, fixing the false-positive rate on real videos to approximately $5\%$. This provides a consistent operating point for comparing different detectors without tuning the classification threshold for accuracy. Average Precision (AP) is computed as the standard area under the precision-recall curve, using the predicted probability of the fake class for ranking.

\section{Dataset Details}
\label{app:dataset_detail}

We conduct experiments on AIGVDBench~\citep{ma2026your}, which contains 31 generator--task configurations, including 20 open-source and 11 closed-source settings. Each open-source setting provides 20,000 generated videos and 20,000 corresponding real videos, where each real video is paired with a generated video under the same prompt.

\paragraph{Training and Evaluation.}
Following the cross-generator evaluation protocol, we select Open-Sora as the sole source of synthetic videos for training. Specifically, we use 14,000 Open-Sora-generated videos and 14,000 matched real videos for training. The remaining evaluation data are not used for model optimization. For the 20 open-source generator--task configurations, we use 3,000 generated videos and 3,000 matched real videos from each configuration for testing. For the 11 closed-source configurations, each setting contains 2,000 generated videos and 2,000 matched real videos for evaluation. Therefore, all generators other than Open-Sora are unseen during training, enabling a direct evaluation of TRACE's cross-generator generalization.

\section{Details of Baselines}
\label{app:baseline_detail}
\textbf{\noindent{CNNSpot\citep{wang2020cnn}.}}
CNNSpot is a widely used image-level detector for identifying AI-generated content, which trains a binary classifier directly on RGB images. We adapt CNNSpot to the video detection setting and follow its original architecture and training strategy, while using the same training data and evaluation protocol as other baselines for a fair comparison.

\textbf{\noindent{UnivFD\citep{ojha2023towards}.}}
UnivFD is a universal synthetic image detector that leverages pretrained visual representations(CLIP\citep{radford2021learning}) to generalize across different generative models. We use the official implementation and pretrained model released by the authors and apply it to sampled video frames, aggregating frame-level predictions to obtain video-level scores.

\textbf{\noindent{NPR\citep{tan2024rethinking}.}}
NPR is a synthetic image detector that exploits artifacts introduced by upsampling operations in CNN-based generative networks to capture generation-related forensic traces. We use the official implementation and follow the recommended preprocessing and evaluation settings.

\textbf{\noindent{DDA\citep{chen2025dual}.}}
DDA is a synthetic image detector that improves generalization through dual data alignment. We follow the official implementation and train DDA using the same training data and evaluation protocol as other baselines.

\textbf{\noindent{D3\citep{yang2025d}.}}
D3 is a deepfake detection method that learns discrepancy representations to distinguish real and AI-generated images. We follow its official implementation and train the model under the same data and evaluation protocol as other baselines for a fair comparison.

\textbf{\noindent{DeMamba\citep{chen2026demamba}.}}
DeMamba is a synthetic video detector that incorporates the Mamba architecture to learn discriminative forensic representations. We follow the official implementation and train DeMamba using the same training data and evaluation protocol as other baselines.

\textbf{\noindent{DeCoF\citep{ma2025detecting}.}}
DeCoF is a synthetic video detection method that exploits frame-level consistency to distinguish real and AI-generated videos. We follow the official implementation and train DeCoF using the same training data and evaluation protocol as other baselines.

\textbf{\noindent{WaveRep\citep{corvi2026seeing}.}}
WaveRep is a synthetic video detection method that exploits wavelet-domain representations to capture forensic cues in both the spatial and frequency domains. We follow the official implementation and train WaveRep using the same training data and evaluation protocol as other baselines.

\textbf{\noindent{ReStraV\citep{interno2026ai}.}}
ReStraV is an AI-generated video detection method that learns perceptually straightened representations to improve detection generalization across unseen generators. We follow the official implementation and train ReStraV using the same training data and evaluation protocol as other baselines.

\textbf{\noindent{Qwen2.5-ViT\citep{li2026preserving}.}}
Qwen2.5-ViT is a vision-based baseline built upon the visual encoder of Qwen2.5-VL. We use the pretrained visual encoder as a frozen feature extractor and train a lightweight binary classification head using the same training data and evaluation protocol as TRACE.

\textbf{\noindent{STALL\citep{ben2026training}.}}
STALL is a training-free AI-generated video detection method that exploits spatial-temporal likelihoods to distinguish real and generated videos. We follow the official implementation and apply STALL using the same preprocessing and evaluation protocol as other baselines.

\textbf{\noindent{V-PVP\citep{cui2026rethinking}.}}
V-PVP is an AI-generated video detection method that leverages pretrained video backbones for discriminative forensic representation learning. We follow the official implementation and train V-PVP using the same training data, video sampling strategy, and evaluation protocol as other baselines.

\section{Velocity Representation Analysis}
\label{app:velocity_analysis}

To investigate whether pretrained generative models inherently encode generation-aware information, we analyze the velocity representations extracted from an untrained Wan2.2-A14B model. Specifically, we compare the velocity responses of real and AI-generated videos at different noise levels without updating any parameters of the generative model. As shown in Fig.~\ref{fig:velocity_rep}, the velocity representations already exhibit a strong discriminative capability between real and AI-generated videos, even when the Wan2.2-A14B backbone is completely frozen and has never been optimized for synthetic video detection. This observation suggests that the velocity field of a pretrained generative model implicitly captures characteristics associated with the underlying video generation process, providing a generation-aware forensic cue beyond conventional appearance-based features.

\noindent
\begin{minipage}[t]{0.50\textwidth}
\vspace{0pt}
We further examine how noise perturbation affects the discriminability of the velocity representations. As the noise level increases, the separation between real and AI-generated videos becomes progressively more pronounced, indicating that noise perturbation exposes additional generation-related differences in the velocity space. Notably, the discriminative performance reaches a peak around timestep $t=100$, after which further increasing the noise level yields only marginal improvements. This saturation suggests that a moderate amount of noise is sufficient to reveal the generation-dependent characteristics encoded in the velocity field, while excessive perturbation provides limited additional forensic information. Based on this observation, we adopt $\mathcal{T}=\{0,25,50,75,100\}$ as the probe timesteps in TRACE.
\end{minipage}\hfill
\begin{minipage}[t]{0.47\textwidth}
\vspace{0pt}
\centering
\setlength{\parskip}{0pt}
\includegraphics[width=\linewidth]{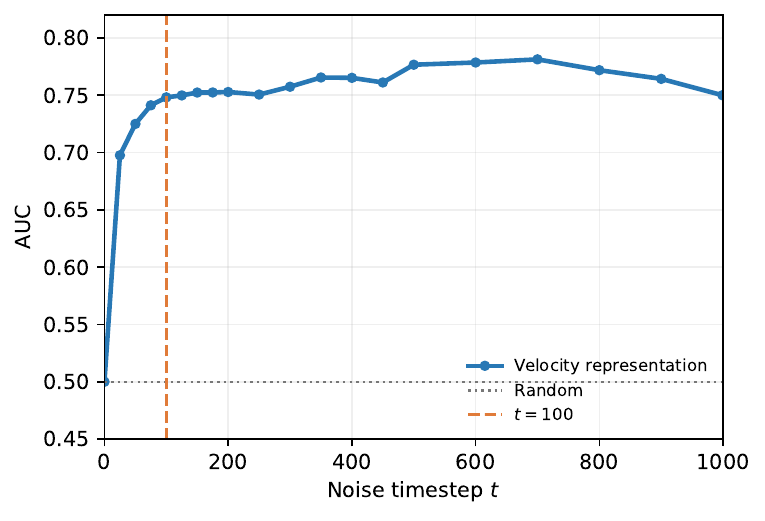}\par
\setlength{\abovecaptionskip}{6pt}
\setlength{\belowcaptionskip}{0pt}
\captionof{figure}{Detection performance based on velocity representations under different noise levels. A relatively low noise level of 100 already yields strong performance, with further noise perturbation providing limited improvement.}
\label{fig:velocity_rep}
\end{minipage}

\par

\section{ACC@5\% FPR Evaluation Results}
\label{app:acc}
\subsection{ACC@5\% FPR on Open-Source Generators}

\begin{table*}[htbp]

  \caption{\textbf{22.59\%} Improvement in ACC@5\%FPR for Open-Source Video Generators.  }
  \resizebox{\textwidth}{!}{
    \begin{tabular}{>{\centering}m{2.5cm}|*{20}{c|}c}
    \toprule
    \multirow{3}{*}{\textbf{Method}} & \multicolumn{6}{c|}{\textbf{I2V}} & \multicolumn{12}{c|}{\textbf{T2V}} & \multicolumn{2}{c|}{\textbf{V2V}} & \multirow{3}{*}{\textbf{AVG}} \\
    \cmidrule(lr){2-7} \cmidrule(lr){8-19} \cmidrule(lr){20-21}
    & \multirow{2}{*}{\makecell[c]{\scriptsize Easy\\\scriptsize Animate}} & \multirow{2}{*}{\makecell[c]{\scriptsize LTX}} & \multirow{2}{*}{\makecell[c]{\scriptsize Pyramid\\\scriptsize Flow}} & \multirow{2}{*}{\makecell[c]{\scriptsize SEINE}} & \multirow{2}{*}{\makecell[c]{\scriptsize SVD}} & \multirow{2}{*}{\makecell[c]{\scriptsize Video\\\scriptsize Crafter}} & \multirow{2}{*}{\makecell[c]{\scriptsize Acc\\\scriptsize Video}} & \multirow{2}{*}{\makecell[c]{\scriptsize Animate\\\scriptsize Diff}} & \multirow{2}{*}{\makecell[c]{\scriptsize Cogvideo\\\scriptsize x1.5}} & \multirow{2}{*}{\makecell[c]{\scriptsize Easy\\\scriptsize Animate}} & \multirow{2}{*}{\makecell[c]{\scriptsize Hunyuan}} & \multirow{2}{*}{\makecell[c]{\scriptsize IPOC}} & \multirow{2}{*}{\makecell[c]{\scriptsize LTX}} & \multirow{2}{*}{\makecell[c]{\scriptsize Open\\\scriptsize Sora}} & \multirow{2}{*}{\makecell[c]{\scriptsize Pyramid\\\scriptsize Flow}} & \multirow{2}{*}{\makecell[c]{\scriptsize Rep\\\scriptsize Video}} & \multirow{2}{*}{\makecell[c]{\scriptsize Video\\\scriptsize Crafter}} & \multirow{2}{*}{\makecell[c]{\scriptsize Wan\\\scriptsize 2.1}} & \multirow{2}{*}{\makecell[c]{\scriptsize Cogvideo\\\scriptsize x1.5}} & \multirow{2}{*}{\makecell[c]{\scriptsize LTX}} & \\
    & & & & & & & & & & & & & & & & & & & & & \\

    \midrule
    
    \multicolumn{22}{c}{\textbf{AI-Generated Image Detection Models}} \\
    \midrule
        CNNSpot & 61.65 & 61.43 & 76.35 & 78.65 & 91.62 & 52.00 & 54.42 & 49.38 & 65.93 & 55.78 & 57.33 & 73.57 & 86.85 & \cellcolor{second}\underline{97.48} & 89.25 & 64.30 & 61.25 & 53.47 & 59.20 & 57.42 & 67.37 \\
            UnivFD & 61.70 & 67.80 & 73.82 & 92.10 & 87.57 & 54.23 & 52.45 & 64.25 & 73.33 & 54.80 & 52.67 & 72.63 & 76.12 & \cellcolor{second}\underline{97.48} & 82.90 & 65.25 & 67.53 & 50.63 & 67.87 & 61.88 & 68.85 \\
        NPR &62.75 & \cellcolor{second}\underline{76.82} & \cellcolor{second}\underline{87.33} & 83.27 & \cellcolor{second}\underline{92.90} & 51.58 & 51.62 & 51.28 & 70.42 & 51.80 & 56.12 & 70.08 & \cellcolor{second}\underline{90.15} & \cellcolor{best}\textbf{97.50} & 84.77 & 62.48 & 57.30 & 50.85 & \cellcolor{second}\underline{74.67} & \cellcolor{second}\underline{70.38} & 69.70 \\
            DDA & 51.10 & 50.63 & 56.08 & 57.60 & 62.02 & 50.05 & 55.30 & 52.67 & 56.80 & 58.45 & 55.87 & 53.08 & 55.27 & \cellcolor{best}\textbf{97.50} & 69.23 & 52.43 & 53.05 & 52.93 & 50.95 & 50.83 & 57.09 \\
    D3 &63.45 & 68.10 & 75.88 & \cellcolor{second}\underline{93.22} & 91.92 & 63.45 & 53.17 & 59.67 & 72.52 & 58.52 & 53.43 & 76.37 & 77.57 & \cellcolor{best}\textbf{97.50} & 88.05 & 67.30 & 72.22 & 52.63 & 65.78 & 62.53 & 70.66 \\

    \midrule
    
    \multicolumn{22}{c}{\textbf{AI-Generated Video Detection Models}} \\
    \midrule
    DeMamba &57.62 & 57.33 & 65.28 & 70.52 & 83.35 & 50.57 & 53.35 & 56.77 & 65.05 & 59.80 & 56.87 & 65.13 & 70.35 & \cellcolor{best}\textbf{97.50} & 77.30 & 57.17 & 76.93 & 51.62 & 57.37 & 54.68 & 64.23 \\
    DeCoF & 60.02 & 64.70 & 70.02 & 80.62 & 90.82 & 52.32 & 60.48 & 59.50 & \cellcolor{second}\underline{77.63} & 63.38 & \cellcolor{second}\underline{61.28} & \cellcolor{second}\underline{84.98} & 81.25 & \cellcolor{best}\textbf{97.50} & \cellcolor{second}\underline{93.65} & 75.95 & 76.75 & 55.32 & 64.67 & 62.85 & \cellcolor{second}\underline{71.68} \\
    WaveRep & 49.93 & 49.85 & 49.97 & 50.22 & 50.20 & 49.77 & 49.65 & 49.90 & 50.05 & 49.95 & 48.98 & 49.65 & 50.27 & 49.70 & 50.13 & 49.85 & 50.57 & 49.72 & 49.33 & 49.78 & 49.87 \\
    ReStraV & \cellcolor{second}\underline{91.87} & 49.02 & 49.17 & 47.62 & 48.12 & 47.70 & 48.73 & 47.52 & 59.55 & \cellcolor{second}\underline{85.12} & 48.38 & 80.80 & 51.02 & 94.60 & 48.83 & \cellcolor{second}\underline{86.37} & 47.52 & \cellcolor{second}\underline{63.75} & 51.27 & 49.57 & 59.83 \\
    Qwen2.5-ViT & 49.70 & 50.20 & 51.05 & 50.27 & 50.08 & 50.93 & 52.03 & 51.88 & 53.92 & 50.50 & 52.02 & 53.28 & 51.28 & 53.92 & 51.70 & 51.55 & 51.70 & 50.47 & 53.28 & 50.43 & 51.51 \\
    STALL & 50.65 & 49.77 & 49.65 & 49.30 & 51.57 & 52.22 & 53.62 & 55.93 & 53.00 & 55.67 & 53.78 & 56.97 & 49.48 & 53.18 & 52.92 & 51.67 & 50.77 & 53.17 & 50.60 & 49.83 & 52.19 \\
    V-PVP & 57.55 & 61.05 & 68.17 & 74.60 & 81.32 & \cellcolor{second}\underline{68.85} & \cellcolor{second}\underline{62.43} & \cellcolor{second}\underline{71.75} & 63.42 & 62.85 & 61.08 & 70.30 & 66.68 & 97.35 & 81.33 & 64.92 & \cellcolor{second}\underline{85.70} & 57.43 & 58.90 & 57.75 & 68.67 \\
    \textbf{\textit{TRACE} } & \cellcolor{best}\textbf{93.98} & \cellcolor{best}\textbf{95.58} & \cellcolor{best}\textbf{97.08} & \cellcolor{best}\textbf{97.47} & \cellcolor{best}\textbf{97.47} & \cellcolor{best}\textbf{97.47} & \cellcolor{best}\textbf{91.13} & \cellcolor{best}\textbf{97.47} & \cellcolor{best}\textbf{92.38} & \cellcolor{best}\textbf{95.85} & \cellcolor{best}\textbf{92.12} & \cellcolor{best}\textbf{94.92} & \cellcolor{best}\textbf{96.82} & 97.47 & \cellcolor{best}\textbf{97.42} & \cellcolor{best}\textbf{90.85} & \cellcolor{best}\textbf{97.47} & \cellcolor{best}\textbf{78.08} & \cellcolor{best}\textbf{92.67} & \cellcolor{best}\textbf{91.77} & \cellcolor{best}\textbf{94.27} \\
    \bottomrule
    \end{tabular}%
  }
  \label{tab:open-source-acc}%
\end{table*}%

Table~\ref{tab:open-source-acc} reports ACC@5\% FPR on the open-source generator subset of AIGVDBench. TRACE achieves an average accuracy of \textbf{94.27\%}, substantially exceeding the strongest baseline, DeCoF (71.68\%), by \textbf{22.59} percentage points. TRACE obtains the highest accuracy in 19 of the 20 generator--task configurations; the only exception is the in-domain Open-Sora setting, where TRACE achieves 97.47\%, closely matching the best baseline result of 97.50\%. More importantly, TRACE maintains consistently high accuracy across diverse generator--task configurations despite being trained exclusively on Open-Sora. This consistent performance across unseen generation sources highlights the strong cross-generator generalization of TRACE and suggests that its trajectory-based representation is less dependent on generator-specific artifacts.

\subsection{ACC@5\% FPR on Closed-Source Generators}

\begin{table*}[htbp]
\centering

\caption{\textbf{23.27\%} Improvement in ACC@5\%FPR over Existing Methods on Closed-Source Video Generators.  }

\resizebox{\textwidth}{!}{
\begin{tabular}{
>{\centering\arraybackslash}m{3cm}|*{11}{c|}c
}

\toprule

\multirow{3}{*}{\textbf{Method}} 
& \multicolumn{11}{c|}{\textbf{Closed-Source Approaches}}
& \multirow{3}{*}{\textbf{AVG}} \\

\cmidrule(lr){2-12}

& \makecell[c]{\scriptsize Gen2}
& \makecell[c]{\scriptsize Gen3}
& \makecell[c]{\scriptsize Jimeng}
& \makecell[c]{\scriptsize Luma}
& \makecell[c]{\scriptsize Open\\\scriptsize Sora}
& \makecell[c]{\scriptsize Sora}
& \makecell[c]{\scriptsize Causvid\\\scriptsize 24fps}
& \makecell[c]{\scriptsize Kling}
& \makecell[c]{\scriptsize Pika}
& \makecell[c]{\scriptsize Vidu}
& \makecell[c]{\scriptsize Wan}
\\

\midrule

\multicolumn{13}{c}{\textbf{AI-Generated Image Detection Models}} \\

\midrule
CNNSpot &58.72 & 65.32 & 57.08 & 62.08 & 64.32 & 62.42 & 63.70 & 65.42 & 66.54 & 70.56 & 62.90 & 63.55 \\
UnivFD & 57.66 & 61.40 & 57.76 & 59.58 & 66.56 & 59.62 & 69.70 & 64.92 & 60.66 & 67.62 & 60.48 & 62.36 \\
NPR & 58.68 & 65.78 & 57.04 & 62.58 & 65.32 & 63.48 & 59.40 & 66.98 & 66.80 & 66.52 & 62.04 & 63.15 \\
DDA &  62.34 & 67.70 & 64.40 & 65.68 & 74.04 & 66.00 & 63.86 & 65.60 & 63.60 & 65.64 & 65.90 & 65.89 \\
D3 & 57.82 & 61.30 & 57.80 & 59.08 & 63.12 & 59.20 & 72.82 & 64.28 & 59.78 & 65.12 & 60.64 & 61.91 \\

\midrule
\multicolumn{13}{c}{\textbf{AI-Generated Video Detection Models}} \\

\midrule

DeMamba &62.46 & 66.28 & 59.84 & 63.22 & 73.76 & 63.84 & 64.82 & 63.72 & 68.16 & 67.78 & 61.82 & 65.06 \\
DeCoF& 58.60 & 64.24 & 57.12 & 65.48 & \cellcolor{second}\underline{79.54} & 65.50 & \cellcolor{second}\underline{87.38} & \cellcolor{second}\underline{69.22} & 71.68 & \cellcolor{second}\underline{72.30} & 65.32 & \cellcolor{second}\underline{68.76} \\
WaveRep & 60.24 & 59.60 & 59.28 & 59.10 & 59.76 & 60.32 & 59.48 & 58.84 & 59.46 & 59.90 & 59.26 & 59.57 \\
ReStraV  & 57.06 & 57.50 & 57.04 & 57.46 & 58.86 & 59.16 & 57.06 & 58.63 & 57.34 & 57.96 & 60.64 & 58.06 \\
Qwen2.5-ViT &69.32 & 61.28 & 68.24 & 60.28 & 63.46 & 59.84 & 60.38 & 59.42 & 61.16 & 59.64 & 60.24 & 62.11 \\
STALL &\cellcolor{second}\underline{75.00} & 68.28 & \cellcolor{second}\underline{69.86} & \cellcolor{second}\underline{71.28} & 72.54 & \cellcolor{second}\underline{66.30} & 67.52 & 64.40 & 69.08 & 63.70 & \cellcolor{second}\underline{68.08} & 68.73 \\
V-PVP &67.98 & \cellcolor{second}\underline{69.10} & 61.84 & 64.62 & 78.86 & 64.42 & 62.82 & 67.48 & \cellcolor{second}\underline{82.18} & 67.92 & 66.54 & 68.52 \\
\textbf{\textit{TRACE} } & \cellcolor{best}\textbf{96.06} & \cellcolor{best}\textbf{90.10} & \cellcolor{best}\textbf{96.08} & \cellcolor{best}\textbf{92.92} & \cellcolor{best}\textbf{96.18} & \cellcolor{best}\textbf{83.14} & \cellcolor{best}\textbf{93.26} & \cellcolor{best}\textbf{92.98} & \cellcolor{best}\textbf{96.82} & \cellcolor{best}\textbf{88.16} & \cellcolor{best}\textbf{86.68} & \cellcolor{best}\textbf{92.03} \\

\bottomrule

\end{tabular}
}

\label{tab:closed_source_acc}

\end{table*}

Table~\ref{tab:closed_source_acc} reports ACC@5\% FPR on the closed-source generator subset. TRACE achieves an average accuracy of \textbf{92.03\%}, substantially outperforming DeCoF (68.76\%), the strongest baseline by average accuracy, by \textbf{23.27} percentage points. Notably, TRACE consistently outperforms all compared methods across all 11 generator--task configurations, with accuracy ranging from 83.14\% on Sora to 96.82\% on Pika. Such consistent improvements across diverse closed-source generators demonstrate that TRACE generalizes effectively beyond the Open-Sora training distribution, despite having no access to these generators during training. The results further indicate that trajectory-based representations provide a robust forensic signal for distinguishing AI-generated videos under a strict 5\% false-positive-rate constraint.

\subsection{ACC@5\% FPR under Robustness Tests}
\begin{figure}[htbp]
\centering
\includegraphics[width=\textwidth]{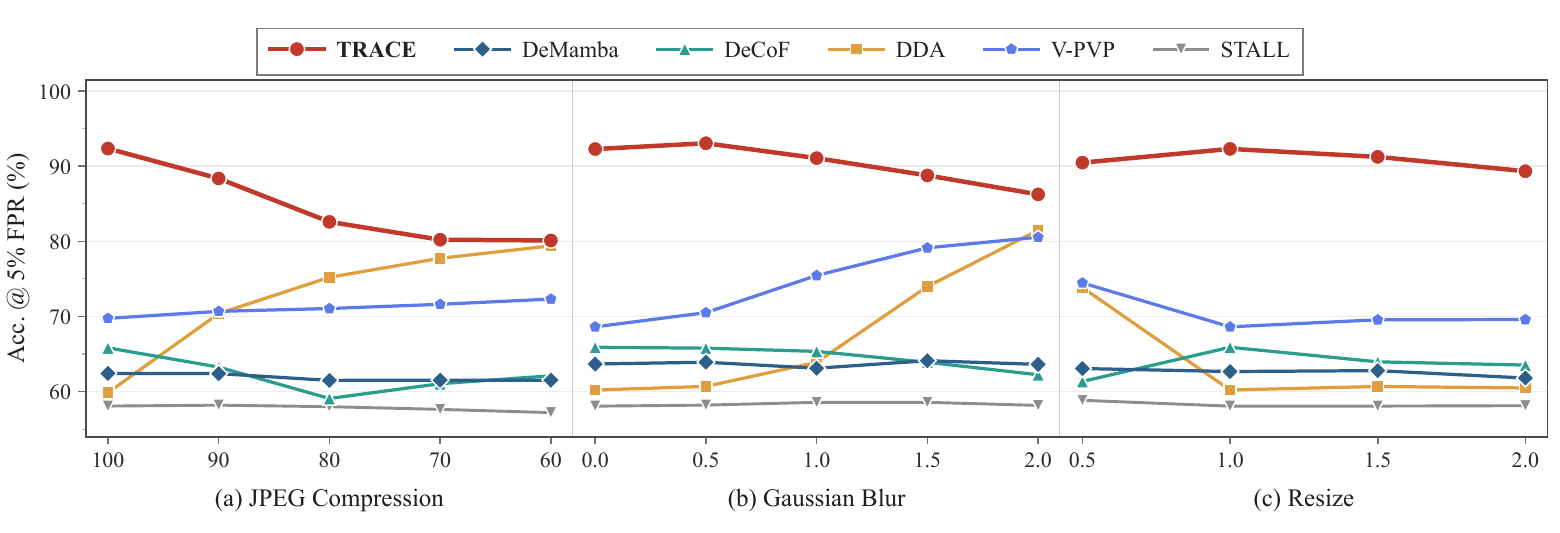}
\caption{Robustness evaluation of TRACE in terms of ACC@5\% FPR under various video distortions, including JPEG compression, Gaussian blur, and resizing.}
\label{fig:robustness_Acc}
\end{figure}

We further evaluate the robustness of TRACE on AIGVDBench under JPEG compression ($q\in\{100,90,80,70,60\}$), Gaussian blur ($\sigma\in\{0.0,0.5,1.0,1.5,2.0\}$), and resolution scaling ($s\in\{0.5,1.0,1.5,2.0\}$, with restoration to the original size). Perturbations are applied to raw video frames prior to model-specific preprocessing. We report ACC@5\% FPR across all 14 settings, as shown in Figure~\ref{fig:robustness_Acc}. TRACE consistently achieves the best performance across all evaluated distortion conditions, maintaining a clear advantage over competing methods under varying levels of compression, blur, and resolution changes. These results demonstrate that TRACE is not only effective across diverse generation sources, but also robust to common distortions that substantially alter the visual appearance of video inputs. The consistent performance under such perturbations further supports the generalizability of trajectory-based forensic representations beyond clean, unaltered video content.

\section{AP Evaluation Results}
\label{app:ap}
\subsection{AP on Open-Source Generators}

\begin{table*}[htbp]

  \caption{\textbf{14.04\%} Improvement in AP for Open-Source Video Generators.   }
  \resizebox{\textwidth}{!}{
    \begin{tabular}{>{\centering}m{2.5cm}|*{20}{c|}c}
    \toprule
    \multirow{3}{*}{\textbf{Method}} & \multicolumn{6}{c|}{\textbf{I2V}} & \multicolumn{12}{c|}{\textbf{T2V}} & \multicolumn{2}{c|}{\textbf{V2V}} & \multirow{3}{*}{\textbf{AVG}} \\
    \cmidrule(lr){2-7} \cmidrule(lr){8-19} \cmidrule(lr){20-21}
    & \multirow{2}{*}{\makecell[c]{\scriptsize Easy\\\scriptsize Animate}} & \multirow{2}{*}{\makecell[c]{\scriptsize LTX}} & \multirow{2}{*}{\makecell[c]{\scriptsize Pyramid\\\scriptsize Flow}} & \multirow{2}{*}{\makecell[c]{\scriptsize SEINE}} & \multirow{2}{*}{\makecell[c]{\scriptsize SVD}} & \multirow{2}{*}{\makecell[c]{\scriptsize Video\\\scriptsize Crafter}} & \multirow{2}{*}{\makecell[c]{\scriptsize Acc\\\scriptsize Video}} & \multirow{2}{*}{\makecell[c]{\scriptsize Animate\\\scriptsize Diff}} & \multirow{2}{*}{\makecell[c]{\scriptsize Cogvideo\\\scriptsize x1.5}} & \multirow{2}{*}{\makecell[c]{\scriptsize Easy\\\scriptsize Animate}} & \multirow{2}{*}{\makecell[c]{\scriptsize Hunyuan}} & \multirow{2}{*}{\makecell[c]{\scriptsize IPOC}} & \multirow{2}{*}{\makecell[c]{\scriptsize LTX}} & \multirow{2}{*}{\makecell[c]{\scriptsize Open\\\scriptsize Sora}} & \multirow{2}{*}{\makecell[c]{\scriptsize Pyramid\\\scriptsize Flow}} & \multirow{2}{*}{\makecell[c]{\scriptsize Rep\\\scriptsize Video}} & \multirow{2}{*}{\makecell[c]{\scriptsize Video\\\scriptsize Crafter}} & \multirow{2}{*}{\makecell[c]{\scriptsize Wan\\\scriptsize 2.1}} & \multirow{2}{*}{\makecell[c]{\scriptsize Cogvideo\\\scriptsize x1.5}} & \multirow{2}{*}{\makecell[c]{\scriptsize LTX}} & \\
    & & & & & & & & & & & & & & & & & & & & & \\

    \midrule
    
    \multicolumn{22}{c}{\textbf{AI-Generated Image Detection Models}} \\
\midrule

CNNSpot 
&76.81 & 75.89 & 89.96 & 91.94 & 97.94 & 56.33 & 60.80 & 44.13 & 80.19 & 64.95 & 69.37 & 87.14 & 95.76 & \cellcolor{best}\textbf{100.0} & 96.69 & 78.23 & 74.66 & 59.14 & 72.33 & 70.71 & 77.15 \\

UnivFD
& 76.02 & 81.60 & 87.90 & 97.33 & 95.62 & 66.40 & 59.31 & 78.20 & 87.76 & 66.26 & 59.08 & 87.07 & 89.72 & \cellcolor{second}\underline{99.99} & 93.81 & 80.11 & 83.62 & 53.36 & 82.74 & 74.47 & 80.02 \\

NPR
& 77.23 & \cellcolor{second}\underline{89.54} & \cellcolor{second}\underline{95.48} & 93.66 & \cellcolor{second}\underline{98.11} & 55.97 & 49.52 & 51.28 & 82.30 & 50.56 & 62.32 & 83.59 & \cellcolor{second}\underline{97.03} & \cellcolor{best}\textbf{100.0} & 94.35 & 73.41 & 70.75 & 49.68 & \cellcolor{second}\underline{88.44} & \cellcolor{second}\underline{85.31} & 77.43 \\

DDA
& 53.99 & 51.82 & 65.55 & 66.08 & 74.58 & 45.96 & 64.30 & 63.08 & 67.66 & 72.05 & 66.42 & 58.64 & 62.33 & \cellcolor{best}\textbf{100.0} & 84.32 & 54.82 & 61.28 & 60.66 & 53.13 & 51.23 & 63.90 \\

D3
& 78.10 & 83.20 & 89.81 & \cellcolor{second}\underline{98.01} & 97.41 & 79.88 & 61.91 & 73.15 & 87.14 & 72.61 & 60.18 & 90.01 & 90.97 & \cellcolor{best}\textbf{100.0} & 96.07 & 82.17 & 88.17 & 58.33 & 81.56 & 77.40 & 82.30 \\

\midrule
\multicolumn{22}{c}{\textbf{AI-Generated Video Detection Models}} \\
\midrule

DeMamba
&70.27 & 69.98 & 81.21 & 87.11 & 94.42 & 55.97 & 61.80 & 71.17 & 80.15 & 75.52 & 70.51 & 80.30 & 85.09 & \cellcolor{best}\textbf{100.0} & 91.52 & 68.23 & 91.58 & 59.10 & 70.70 & 65.44 & 76.50 \\

DeCoF
&75.37 & 80.99 & 86.25 & 92.98 & 97.17 & 58.85 & \cellcolor{second}\underline{76.68} & 73.47 & \cellcolor{second}\underline{91.03} & 80.09 & \cellcolor{second}\underline{77.99} & \cellcolor{second}\underline{94.97} & 93.05 & \cellcolor{best}\textbf{100.0} & \cellcolor{second}\underline{98.39} & 90.29 & 90.84 & 66.73 & 80.61 & 79.31 & \cellcolor{second}\underline{84.25} \\

WaveRep
&  49.91 & 49.17 & 49.74 & 50.18 & 49.76 & 49.59 & 48.94 & 49.32 & 50.41 & 51.00 & 46.62 & 49.54 & 50.94 & 49.88 & 50.90 & 49.68 & 52.60 & 49.38 & 47.83 & 49.29 & 49.73 \\

ReStraV
&\cellcolor{second}\underline{97.52} & 46.12 & 48.40 & 45.03 & 55.46 & 38.47 & 45.11 & 33.18 & 72.13 & \cellcolor{second}\underline{93.27} & 46.11 & 89.05 & 50.30 & 98.69 & 45.02 & \cellcolor{second}\underline{93.71} & 33.06 & \cellcolor{second}\underline{76.96} & 53.15 & 49.10 & 60.49 \\

Qwen2.5-ViT
&50.74 & 49.96 & 53.82 & 51.51 & 51.08 & 51.88 & 56.92 & 53.88 & 61.74 & 52.41 & 57.27 & 59.24 & 53.66 & 61.36 & 55.83 & 55.66 & 56.20 & 52.60 & 58.65 & 50.90 & 54.77 \\

STALL
&51.46 & 47.92 & 48.22 & 46.95 & 52.66 & 55.07 & 59.16 & 66.87 & 57.53 & 64.13 & 60.52 & 65.86 & 47.65 & 60.23 & 57.94 & 54.84 & 57.99 & 58.76 & 51.28 & 48.93 & 55.70 \\

V-PVP
& 70.76 & 75.19 & 84.03 & 89.02 & 92.97 & \cellcolor{second}\underline{83.86} & 76.34 & \cellcolor{second}\underline{86.38} & 77.56 & 78.24 & 75.89 & 85.23 & 81.99 & 99.90 & 93.17 & 78.91 & \cellcolor{second}\underline{94.88} & 70.04 & 72.30 & 70.68 & 81.87 \\

\textbf{\textit{TRACE}}
&\cellcolor{best}\textbf{98.39} & \cellcolor{best}\textbf{99.04} & \cellcolor{best}\textbf{99.69} & \cellcolor{best}\textbf{100.0} & \cellcolor{best}\textbf{100.0} & \cellcolor{best}\textbf{100.0} & \cellcolor{best}\textbf{96.75} & \cellcolor{best}\textbf{99.95} & \cellcolor{best}\textbf{97.51} & \cellcolor{best}\textbf{99.25} & \cellcolor{best}\textbf{97.39} & \cellcolor{best}\textbf{98.88} & \cellcolor{best}\textbf{99.54} & \cellcolor{best}\textbf{100.0} & \cellcolor{best}\textbf{99.94} & \cellcolor{best}\textbf{96.79} & \cellcolor{best}\textbf{100.0} & \cellcolor{best}\textbf{87.56} & \cellcolor{best}\textbf{97.71} & \cellcolor{best}\textbf{97.38} & \cellcolor{best}\textbf{98.29} \\
    
    \bottomrule
    \end{tabular}%
  }
  \label{tab:open-source-AP}%
\end{table*}%

Table~\ref{tab:open-source-AP} reports AP on the open-source generator subset. TRACE achieves an average AP of \textbf{98.29\%}, substantially surpassing the strongest baseline, DeCoF (84.25\%), by \textbf{14.04} percentage points. TRACE ranks first in 19 of the 20 generator--task configurations and ties for first on the in-domain Open-Sora setting. Moreover, TRACE achieves an AP above 96\% in all configurations except Wan2.1 text-to-video, where it still attains 87.56\%. The consistently high AP across diverse generation sources demonstrates that TRACE maintains reliable ranking capability beyond its Open-Sora training distribution, indicating that its trajectory-based representation provides strong and generalizable discrimination between real and AI-generated videos.

\subsection{AP on Closed-Source Generators}

\begin{table*}[htbp]
\centering

\caption{\textbf{25.06\%} Improvement in AP over Existing Methods on Closed-Source Video Generators. }

\resizebox{\textwidth}{!}{
\begin{tabular}{
>{\centering\arraybackslash}m{3cm}|*{11}{c|}c
}

\toprule

\multirow{3}{*}{\textbf{Method}} 
& \multicolumn{11}{c|}{\textbf{Closed-Source Approaches}}
& \multirow{3}{*}{\textbf{AVG}} \\

\cmidrule(lr){2-12}
& \makecell[c]{\scriptsize Gen2}
& \makecell[c]{\scriptsize Gen3}
& \makecell[c]{\scriptsize Jimeng}
& \makecell[c]{\scriptsize Luma}
& \makecell[c]{\scriptsize Open\\\scriptsize Sora}
& \makecell[c]{\scriptsize Sora}
& \makecell[c]{\scriptsize Causvid\\\scriptsize 24fps}
& \makecell[c]{\scriptsize Kling}
& \makecell[c]{\scriptsize Pika}
& \makecell[c]{\scriptsize Vidu}
& \makecell[c]{\scriptsize Wan}

\\

\midrule

\multicolumn{13}{c}{\textbf{AI-Generated Image Detection Models}} \\

\midrule
CNNSpot & 34.59 & 57.87 & 25.27 & 50.38 & 56.53 & 52.12 & 59.84 & 54.12 & 60.46 & 71.40 & 50.41 & 52.09 \\
UnivFD &33.66 & 43.14 & 27.79 & 41.46 & 64.83 & 42.55 & 71.76 & 55.33 & 46.37 & 66.24 & 45.75 & 48.99 \\
NPR &30.31 & 52.14 & 28.97 & 48.46 & 57.79 & 49.64 & 35.55 & 55.81 & 61.10 & 64.36 & 44.63 & 48.07 \\
DDA & 54.58 & 67.46 & 56.80 & 61.17 & 78.01 & 64.11 & 58.89 & 62.29 & 57.29 & 63.70 & 62.31 & 62.42 \\
D3 & 31.79 & 42.31 & 30.32 & 36.73 & 57.56 & 37.42 & 75.67 & 54.76 & 43.70 & 59.13 & 43.77 & 46.65 \\

\midrule
\multicolumn{13}{c}{\textbf{AI-Generated Video Detection Models}} \\

\midrule

DeMamba & 56.91 & 61.08 & 43.76 & 53.94 & 76.28 & 57.60 & 60.66 & 51.19 & 69.54 & 66.63 & 50.33 & 58.90 \\
DeCoF& 41.54 & 60.29 & 28.56 & 62.99 & \cellcolor{second}\underline{86.18} & 64.26 & \cellcolor{second}\underline{92.76} & \cellcolor{second}\underline{69.18} & 76.47 & \cellcolor{second}\underline{77.43} & 62.21 & 65.63 \\
WaveRep &45.68 & 42.92 & 42.41 & 41.21 & 44.24 & 46.42 & 44.09 & 38.84 & 42.97 & 45.01 & 41.91 & 43.25 \\
ReStraV  & 26.42 & 32.81 & 25.62 & 31.40 & 35.66 & 44.26 & 32.75 & 34.44 & 32.27 & 35.71 & 45.70 & 34.28 \\
Qwen2.5-ViT & 68.06 & 48.56 & 66.41 & 45.97 & 56.16 & 43.18 & 46.80 & 41.13 & 49.05 & 44.35 & 44.27 & 50.36 \\
STALL  & \cellcolor{second}\underline{80.45} & \cellcolor{second}\underline{70.48} & \cellcolor{second}\underline{71.19} & \cellcolor{second}\underline{74.80} & 76.58 & \cellcolor{second}\underline{66.22} & 70.92 & 61.18 & 71.60 & 60.45 & \cellcolor{second}\underline{68.25} & \cellcolor{second}\underline{70.19} \\

V-PVP & 67.67 & 69.55 & 52.73 & 60.48 & 84.66 & 60.69 & 59.17 & 65.71 & \cellcolor{second}\underline{87.68} & 68.52 & 63.40 & 67.30 \\
\textbf{\textit{TRACE} } & \cellcolor{best}\textbf{99.22} & \cellcolor{best}\textbf{93.62} & \cellcolor{best}\textbf{99.25} & \cellcolor{best}\textbf{96.78} & \cellcolor{best}\textbf{99.23} & \cellcolor{best}\textbf{84.22} & \cellcolor{best}\textbf{96.99} & \cellcolor{best}\textbf{96.98} & \cellcolor{best}\textbf{99.83} & \cellcolor{best}\textbf{92.47} & \cellcolor{best}\textbf{89.18} & \cellcolor{best}\textbf{95.25} \\

\bottomrule

\end{tabular}
}

\label{tab:closed_source_AP}

\end{table*}

Table~\ref{tab:closed_source_AP} reports AP on the closed-source generator subset. TRACE achieves an average AP of \textbf{95.25\%}, substantially exceeding the strongest baseline, STALL (70.19\%), by \textbf{25.06} percentage points. TRACE leads in all 11 generator--task configurations, reaching APs of 99.83\% on Pika, 99.25\% on Jimeng, and 99.22\% on Gen2. Although Sora remains the most challenging setting, TRACE still achieves an AP of 84.22\%, substantially higher than the best baseline at 66.22\%. Notably, these closed-source generators are unseen during training, yet TRACE maintains consistently strong ranking performance across all evaluated settings. This result further highlights the strong cross-generator generalization of trajectory-based forensic representations, which remain effective even when the evaluated generators are entirely outside the training distribution.

\begin{figure}[htbp]
\centering
\includegraphics[width=\textwidth]{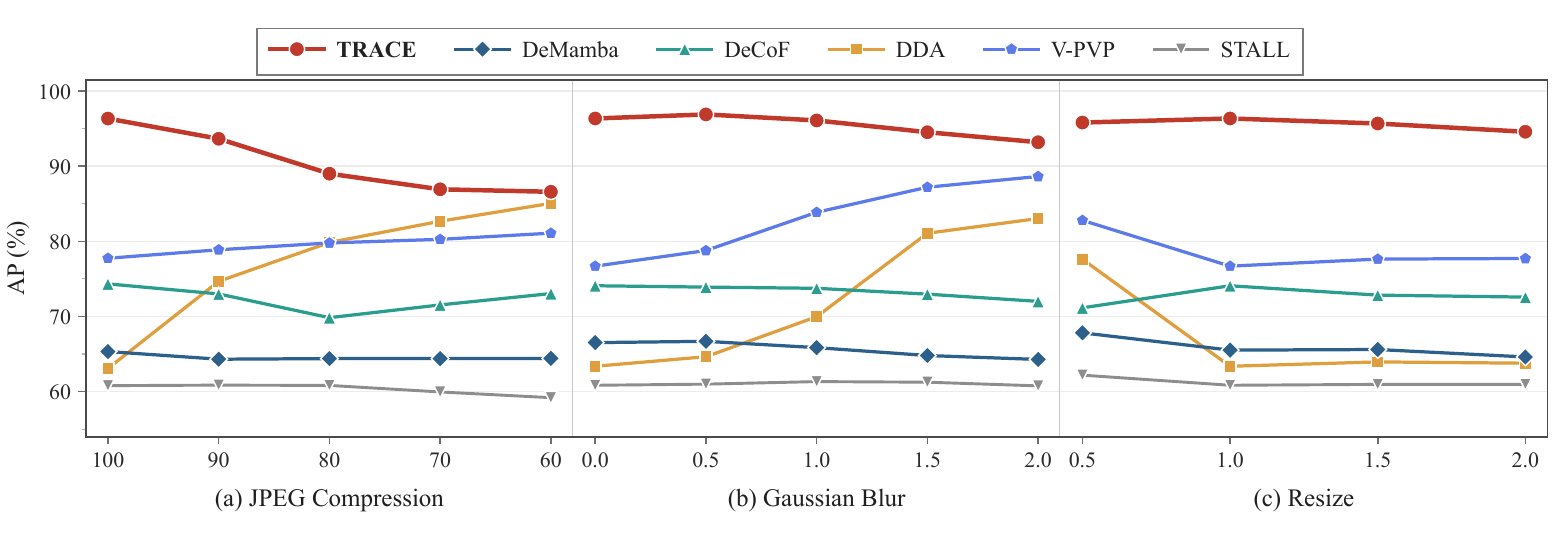}
\caption{Robustness evaluation of TRACE in terms of AP under different video distortions, including JPEG compression, Gaussian blur, and resizing.}
\label{fig:robustness_AP}
\end{figure}
\subsection{AP under Robustness Tests}

We further evaluate the robustness of TRACE on AIGVDBench under JPEG compression ($q\in\{100,90,80,70,60\}$), Gaussian blur ($\sigma\in\{0.0,0.5,1.0,1.5,2.0\}$), and resolution scaling ($s\in\{0.5,1.0,1.5,2.0\}$, with restoration to the original size). Perturbations are applied to raw video frames prior to model-specific preprocessing. We report AP across all 14 settings, as shown in Figure~\ref{fig:robustness_AP}. TRACE consistently achieves the best AP across all evaluated distortion conditions and maintains a clear advantage over competing methods under different perturbation strengths. This consistent ranking performance indicates that TRACE is robust to substantial changes in the visual appearance of video inputs and does not rely heavily on fragile, distortion-sensitive artifacts. More importantly, the stability of its performance across diverse distortions further demonstrates the generalizability of trajectory-based forensic representations beyond clean video inputs.

\section{Visualization Analysis}
\label{sec:Visualization}

To better understand the forensic evidence exploited by TRACE, we conduct a gradient-based saliency analysis to investigate the regions contributing to its predictions. This visualization provides qualitative insights into whether TRACE relies primarily on global semantic content or localized, generation-related forensic cues.

\subsection{Gradient-based Saliency Analysis}

We first visualize the gradient-based saliency maps of TRACE to examine the spatial regions that contribute most to its forensic predictions. Given the predicted fake score $f(\mathbf{x})$, we compute the gradient of the score with respect to the input representation and aggregate the gradient magnitude over the feature channels to obtain a saliency map:
\begin{equation}
\mathcal{S}(\mathbf{x})
=
\left\|
\frac{\partial f(\mathbf{x})}
{\partial \mathbf{x}}
\right\|.
\end{equation}

\begin{figure}[t]
\centering
\includegraphics[width=\textwidth]{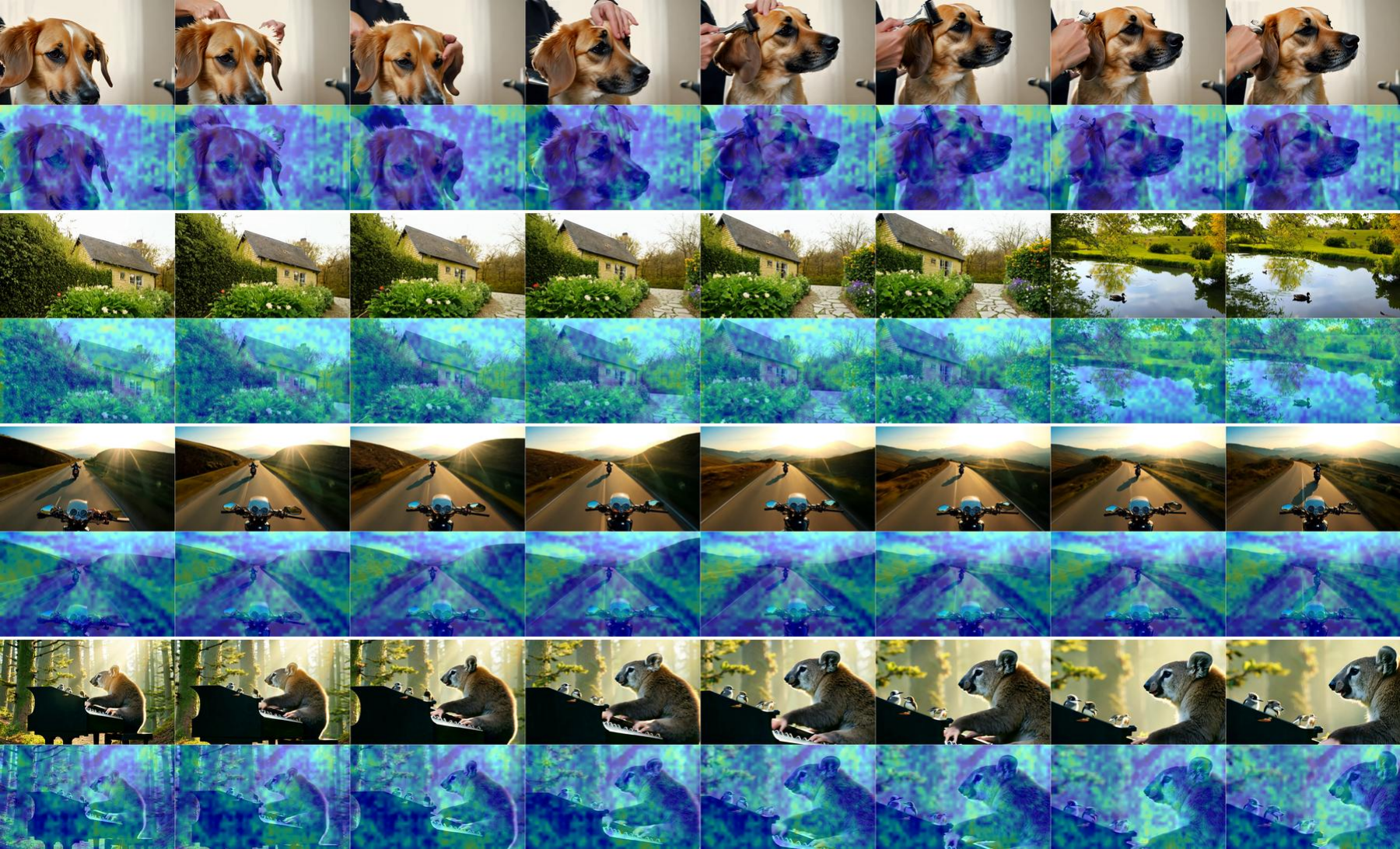}
\caption{\textbf{Saliency analysis.} Grad-CAM visualizations of TRACE at the probe timestep $t{=}50$ on AI-generated videos. The top row shows the original frames, while the bottom row presents the corresponding saliency overlays.}
\label{fig:saliency4_t50_panel}
\end{figure}

The resulting saliency maps are normalized and projected onto the corresponding video frames for visualization. As shown in Figure~\ref{fig:saliency4_t50_panel}, TRACE primarily focuses on localized regions rather than uniformly attending to the entire frame. These highlighted regions may contain fine-grained generation-related artifacts or inconsistencies that are difficult to identify from high-level semantic information alone.

Furthermore, the highlighted regions exhibit a degree of spatial continuity across neighboring frames. This observation suggests that TRACE captures structured forensic cues associated with the evolution of video content over time, rather than relying solely on isolated spatial artifacts. Such behavior is consistent with the design of Trajectory Consistency Estimation, which models frame-to-frame differences in velocity representations under shared probe conditions. Overall, the visualization provides qualitative evidence that TRACE exploits localized and temporally structured cues for AI-generated video detection.

\subsection{Occlusion Analysis}

To further assess the functional importance of the regions highlighted by the saliency maps, we conduct an occlusion-based analysis. Unlike gradient-based visualization, which measures local prediction sensitivity, occlusion analysis directly evaluates the effect of removing specific input regions on the model's prediction. For an input $\mathbf{x}$ and its occluded version $\mathbf{x}^{(r)}$, where region $r$ is replaced with a neutral value, we define the occlusion importance as
\begin{equation}
\Delta_r
=
f(\mathbf{x})-f(\mathbf{x}^{(r)}).
\end{equation}

\begin{figure}[t]
\centering
\includegraphics[width=\textwidth]{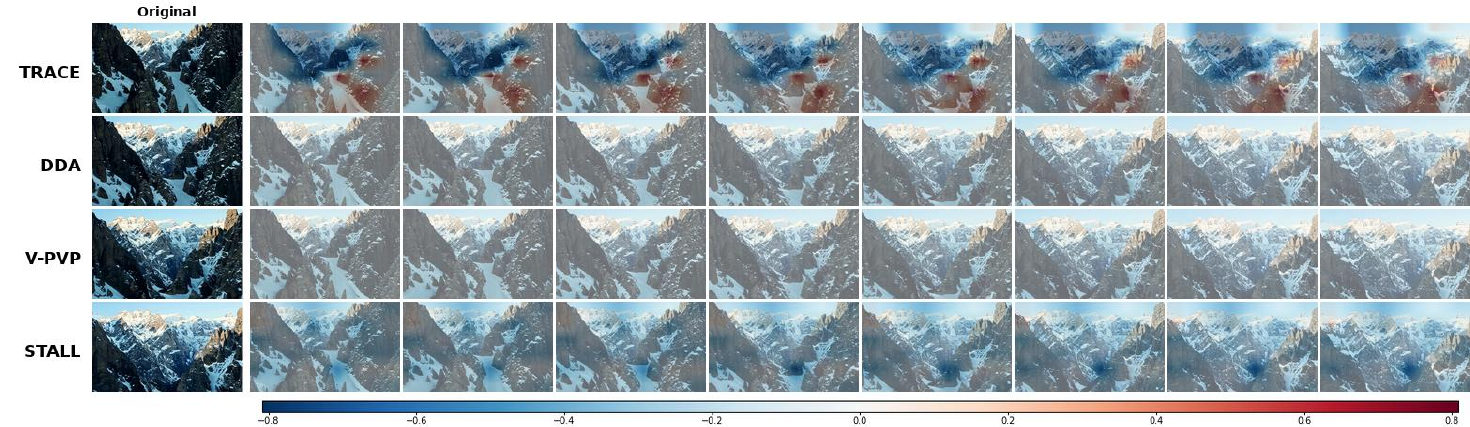}
\caption{\textbf{Occlusion sensitivity analysis.} Patch-occlusion sensitivity maps of TRACE, DDA, V-PVP, and STALL on AI-generated videos. The left column shows the original frames, while the right column presents the corresponding occlusion overlays, where $\Delta=\ell_{\mathrm{orig}}-\ell_{\mathrm{masked}}$ denotes the change in the predicted fake logit after patch masking. All methods share the same color scale.}
\label{fig:occ_1}
\end{figure}

\begin{figure}[t]
\centering
\includegraphics[width=\textwidth]{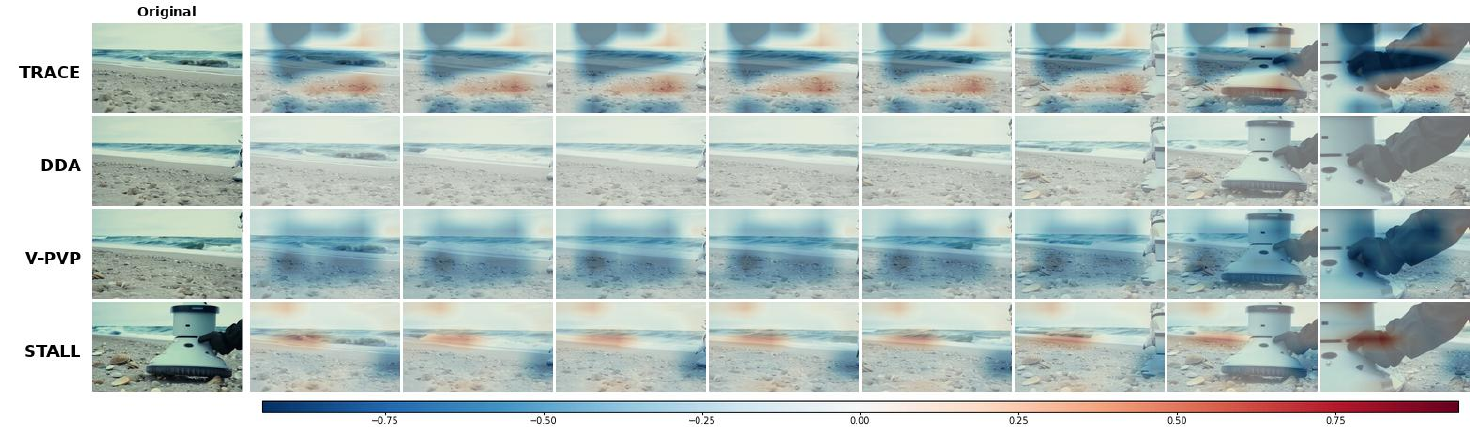}
\caption{\textbf{Occlusion sensitivity analysis.} Patch-occlusion sensitivity maps of TRACE, DDA, V-PVP, and STALL on AI-generated videos. The left column shows the original frames, while the right column presents the corresponding occlusion overlays, where $\Delta=\ell_{\mathrm{orig}}-\ell_{\mathrm{masked}}$ denotes the change in the predicted fake logit after patch masking. All methods share the same color scale.}
\label{fig:occ_2}
\end{figure}

\begin{figure}[t]
\centering
\includegraphics[width=\textwidth]{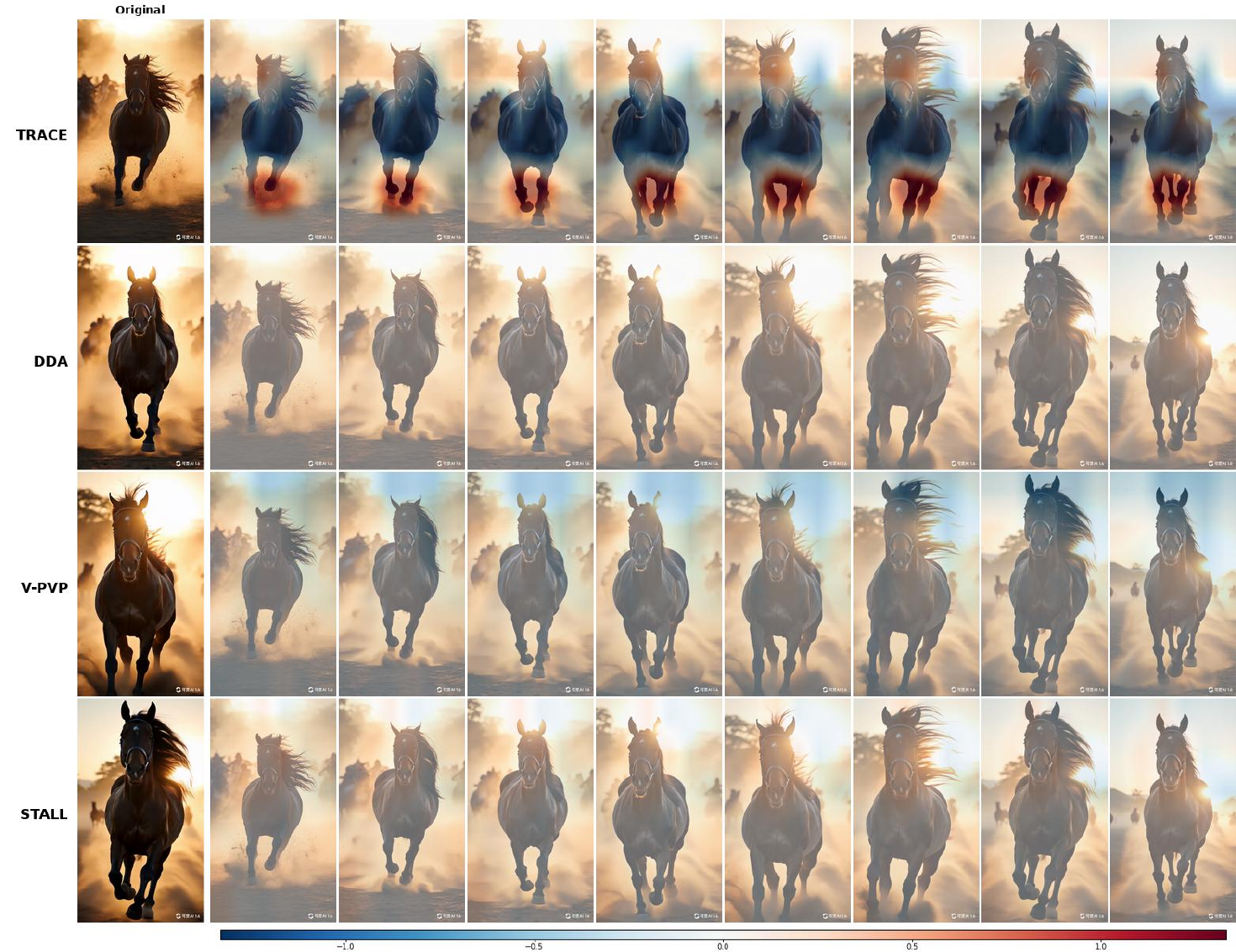}
\caption{\textbf{Occlusion sensitivity analysis.} Patch-occlusion sensitivity maps of TRACE, DDA, V-PVP, and STALL on AI-generated videos. The left column shows the original frames, while the right column presents the corresponding occlusion overlays, where $\Delta=\ell_{\mathrm{orig}}-\ell_{\mathrm{masked}}$ denotes the change in the predicted fake logit after patch masking. All methods share the same color scale.}
\label{fig:occ_3}
\end{figure}

A larger $\Delta_r$ indicates that masking region $r$ causes a greater decrease in the predicted fake score, suggesting that the region makes a stronger contribution to the model's decision. Figures~\ref{fig:occ_1}, \ref{fig:occ_2} and \ref{fig:occ_3} present occlusion sensitivity maps for three representative video examples, comparing TRACE with several competing methods. Across these examples, TRACE exhibits stronger and more spatially concentrated responses to specific patches, indicating that its predictions are particularly sensitive to localized regions containing informative forensic cues. Notably, the highlighted regions frequently correspond to motion-related structures and locally varying content across neighboring frames. This observation suggests that TRACE is sensitive to temporal changes and motion-related inconsistencies, rather than relying exclusively on static appearance artifacts.

Moreover, occluding the highlighted regions produces a more pronounced change in TRACE's prediction, whereas masking less relevant regions generally leads to smaller variations in the fake score. These results provide intervention-based evidence that the regions identified by TRACE are functionally involved in its forensic decisions, rather than merely exhibiting high gradient sensitivity.

Taken together, the saliency and occlusion analyses offer complementary evidence for the learned forensic representations. The former identifies regions that are sensitive to the model's prediction, while the latter verifies their functional contribution through direct input perturbation. Their agreement suggests that TRACE captures localized and structured forensic cues, including cues related to temporal content variations, beyond broad semantic information alone.

\section{t-SNE Visualization}
\label{sec:t-sne}
\begin{figure}[htbp]
    \centering
    \includegraphics[width=\linewidth]{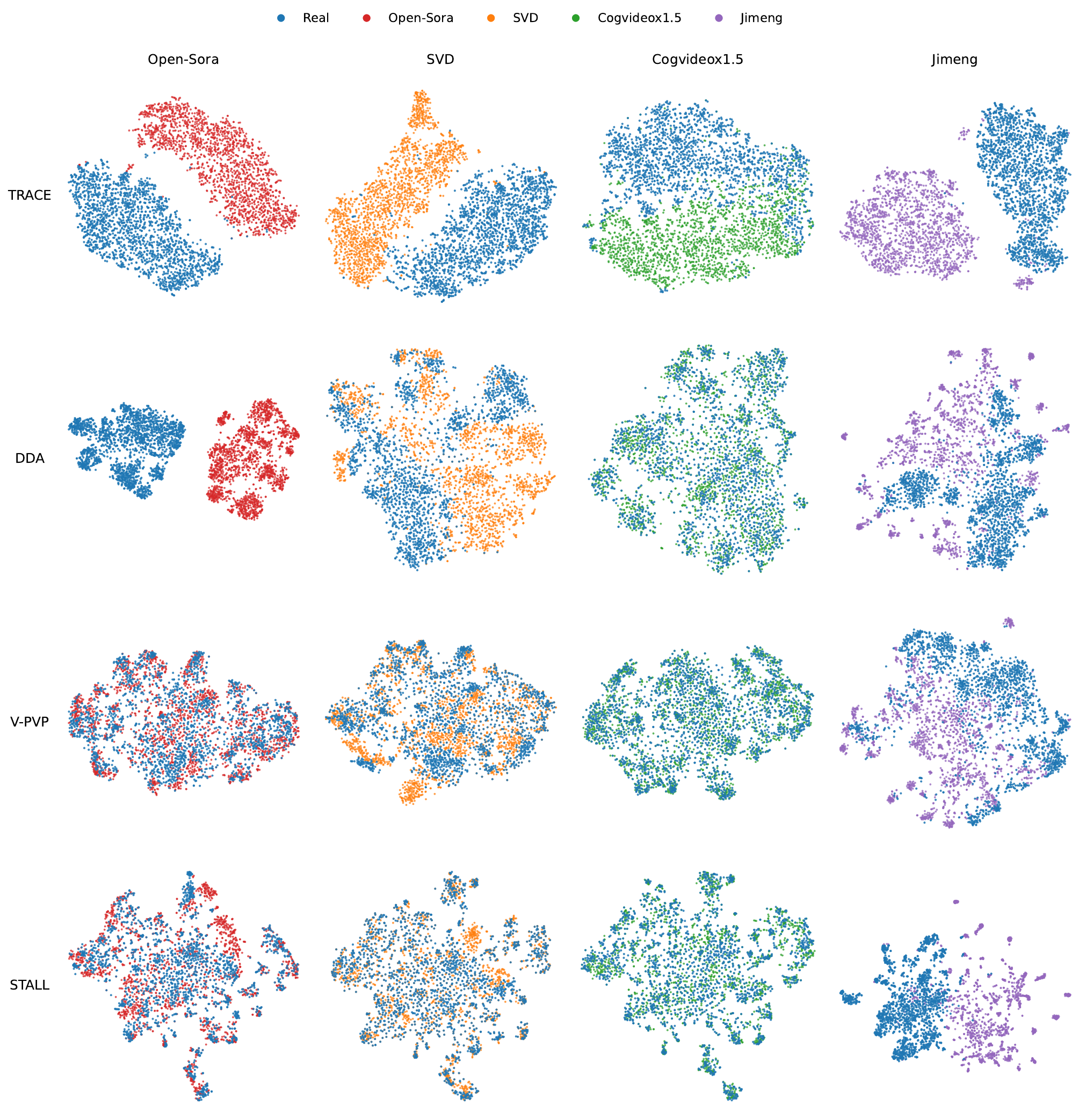}
    \caption{\textbf{t-SNE comparison of detector representations on AIGVDBench.} Rows correspond to TRACE, DDA, V-PVP, and STALL, while columns represent Open-Sora~(T2V), SVD~(I2V), CogVideoX1.5~(V2V), and Jimeng~(closed-source). Each panel visualizes the representations of 2{,}000 real test videos (blue) and 2{,}000 generated test videos (colored) from the corresponding source. TRACE consistently achieves clearer real--fake separation across different generator types, whereas prior methods exhibit better separation mainly on sources aligned with their training distributions and substantial overlap on unseen sources.}
    \label{fig:tsne-2x2}
\end{figure}

Figure~\ref{fig:tsne-2x2} visualizes the fused trajectory features of TRACE and competing methods across four datasets, resulting in 16 t-SNE projections. TRACE exhibits more consistent real--generated separation across datasets, indicating its ability to learn generalizable generation-aware representations. In contrast, competing methods show clear separation only on a subset of datasets or those closely related to their training distributions, while exhibiting greater overlap on other datasets. These visualizations provide qualitative evidence of the stronger cross-dataset generalization of TRACE's learned representations.

\end{document}